\documentclass[11pt]{article}
\usepackage{array}
\usepackage{tabularx}

\usepackage[final]{acl}

\usepackage{microtype}
\usepackage{graphicx}
\usepackage{subcaption}
\usepackage{booktabs} 
\usepackage{lineno} 
\usepackage{xurl}
\usepackage{hyperref}

\usepackage{amsmath}
\usepackage{amssymb}
\usepackage{mathtools}
\usepackage{amsthm}

\usepackage[capitalize,noabbrev]{cleveref}

\theoremstyle{plain}

\theoremstyle{definition}

\theoremstyle{remark}

\usepackage[textsize=tiny]{todonotes}

\usepackage{booktabs} 
\usepackage{multirow}  
\usepackage{tabularx} 
\usepackage{makecell} 
\usepackage{xcolor}
\usepackage[most]{tcolorbox}
\usepackage{nicematrix} 
\usepackage{booktabs}   
\usepackage{array}      
\usepackage{enumitem}
\usepackage{graphicx}
\usepackage[table]{xcolor}

\usepackage{xcolor}
\usepackage{tcolorbox}
\usepackage{listings}
\tcbuselibrary{skins, breakable}
\usepackage{listings}
\usepackage[utf8]{inputenc}
\usepackage{csquotes}
\usepackage{stfloats}
\usepackage{varwidth}
\usepackage{pifont}    
\usepackage{booktabs} 
\usepackage{makecell}  
\usepackage{multirow}

\usepackage{times}
\usepackage{latexsym}

\usepackage[T1]{fontenc}

\usepackage[utf8]{inputenc}

\usepackage{microtype}

\usepackage{inconsolata}

\usepackage{graphicx}

\definecolor{darkgreen}{RGB}{30, 150, 30}

\newcommand{\gcheck}{\textcolor{darkgreen}{\ding{51}}}
\newcommand{\rcross}{\textcolor{red}{\ding{55}}}

\definecolor{bg_gray}{RGB}{245, 245, 245}
\definecolor{prompt_blue}{RGB}{0, 0, 128}
\definecolor{resp_green}{RGB}{0, 100, 0}
\newcommand{\newadd}{}
\newcommand{\newaddd}{}
\newcolumntype{C}[1]{>{\centering\arraybackslash}m{#1}}
\newcolumntype{E}[1]{>{\footnotesize\raggedright\arraybackslash}p{#1}}

\newcommand{\revise}[1]{\textcolor{black}{#1}}

\title{TimeBlind: A Spatio-Temporal Compositionality Benchmark \newline for Video LLMs}

\author{
  \textbf{Baiqi Li\textsuperscript{1}}\thanks{Corresponding author: \href{mailto:baiqili@cs.unc.edu}{baiqili@cs.unc.edu}},
  \textbf{Kangyi Zhao\textsuperscript{2}},
  \textbf{Ce Zhang\textsuperscript{1}},
  \textbf{Chancharik Mitra\textsuperscript{3}}
\\
  \textbf{Jean de Dieu Nyandwi\textsuperscript{3}},
  \textbf{Gedas Bertasius\textsuperscript{1}}
\\
\\
  \textsuperscript{1}University of North Carolina at Chapel Hill \\
  \textsuperscript{2}University of Pittsburgh \\
  \textsuperscript{3}Carnegie Mellon University
}

\begin{document}
\maketitle

\begin{abstract}

Fine-grained spatio-temporal understanding is essential for video reasoning and embodied AI. Yet, while Multimodal Large Language Models (MLLMs) master static semantics, their grasp of temporal dynamics remains brittle. We present TimeBlind, a diagnostic benchmark for compositional spatio-temporal understanding. Inspired by cognitive science, TimeBlind categorizes fine-grained temporal understanding into three levels: recognizing atomic events, characterizing event properties, and reasoning about event interdependencies. Unlike benchmarks that conflate recognition with temporal reasoning, TimeBlind leverages a minimal-pairs paradigm: video pairs share identical static visual content but differ solely in temporal structure, utilizing complementary questions to neutralize language priors. Evaluating over 20 state-of-the-art MLLMs (e.g., GPT-5, Gemini 3 Pro) on 600 curated instances (2400 video-question pairs) reveals that the Instance Accuracy (correctly distinguishing both videos in a pair) of the best performing MLLM is only 48.2\%, far below human performance (98.2\%). These results demonstrate that even \revise{frontier models exhibit significant deficiencies in temporal reasoning}, positioning TimeBlind as a vital diagnostic tool for next-generation video understanding. Dataset and code are available at \url{https://baiqi-li.github.io/timeblind_project/}.

\end{abstract}

\section{Introduction}

\begin{figure}[t]
  \centering
  \includegraphics[width=1\linewidth]{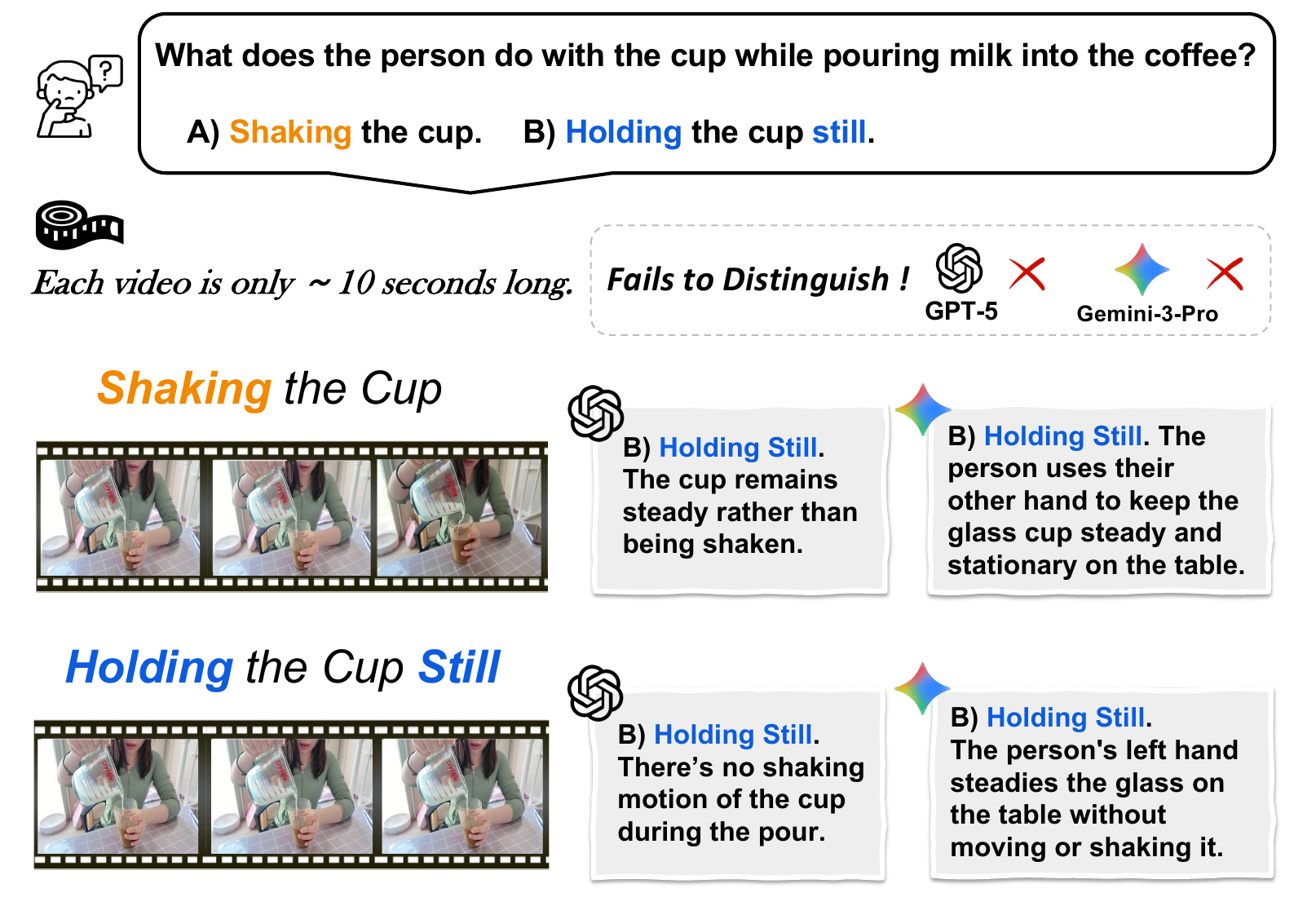}
  \caption{\textbf{An example video pair that shares identical static visual content but differs solely in motion dynamics.} The top video shows a person shaking a cup, while the bottom video shows them holding it still. Even the most advanced models like GPT-5 and Gemini 3 Pro fail to distinguish the actions in the video pair.}
  \label{fig:teaser}
\end{figure}

\begin{table*}[t]
\centering
\caption{\textbf{Comparison with prior temporal video benchmarks.} \newadd{We compare the number of video-question pairs}, coverage of temporal reasoning categories and anti-shortcut design features. TimeBlind uniquely covers all 13 Allen temporal relations\textsuperscript{\dag}, includes causal and comparative reasoning, and employs both paired videos and complementary questions to eliminate static and linguistic shortcuts. The rightmost column shows the gap between human and best model performance, indicating benchmark difficulty.}
\label{tab:comparison}
\renewcommand{\arraystretch}{1.2}
\resizebox{1\textwidth}{!}{
    \begin{tabular}{l c ccccc cc c}
    \toprule
    & & \multicolumn{5}{c}{\textsc{Temporal Reasoning Coverage}} & \multicolumn{2}{c}{\textsc{Anti-Shortcut Design}} & \\
    \cmidrule(lr){3-7} \cmidrule(lr){8-9}
    
    \textsc{Benchmark} & 
    \textsc{Size} & 
    \makecell{\textsc{Event} \\ \textsc{Types}} &
    \makecell{\textsc{Event} \\ \textsc{Attributes}} & 
    \makecell{\textsc{Temporal} \\ \textsc{Topologies}} & 
    \makecell{\textsc{Causal}} & 
    \makecell{\textsc{Cross-Event} \\ \textsc{Comparison}} & 
    \makecell{\textsc{Video} \\ \textsc{Pairs}} &
    \makecell{\textsc{Compl.} \\ \textsc{Questions}} &
    \makecell{\textsc{Human--Model} \\ \textsc{Gap}} \\
    \midrule
    MVBench~\cite{li2024mvbench}         & 4K   & 2 & 2 & 2  & \rcross & \rcross & \rcross & \rcross & -- \\
    TOMATO~\cite{shangguan2024tomato}                 & 1.4K & 2 & 4 & 2  & \rcross & \rcross & \rcross & \rcross & 57.3\% \\
    Vinoground~\cite{zhang2024vinoground}         & 2K   & 1 & 1 & 2  & \rcross & \rcross & \gcheck & \rcross & 40.0\% \\
    TempCompass~\cite{liu2024tempcompass}       & 4K   & 2 & 2 & 2  & \rcross & \gcheck & \gcheck & \rcross & -- \\
    \midrule
    \textbf{TimeBlind (Ours)}            & 2.4K & 2 & 6 & 13\textsuperscript{\dag} & \gcheck & \gcheck & \gcheck & \gcheck & 50.0\% \\
    \bottomrule
    \end{tabular}
}
\vspace{2pt}
{\raggedright \footnotesize \textsuperscript{\dag}Covers all 13 event relations in Allen's Interval Algebra~\cite{allen1983maintaining}: \emph{before, after, meets, met-by, overlaps, overlapped-by, starts, started-by, finishes, finished-by, during, contains, equals}.\par}
\end{table*}

Fine-grained spatio-temporal understanding is fundamental for long-horizon video reasoning~\cite{wang2025lvbench, pan2025basket} and embodied AI~\cite{yang2025cambrian, li2026watchact}. Beyond simply recognizing \emph{what} is present, an intelligent system must infer \emph{what changes}, \emph{how it changes}, and \emph{how multiple changes compose} into causal structures. While recent Multimodal Large Language Models (MLLMs)~\cite{gemini, openai2025gpt5systemcard, bai2025qwen3vltechnicalreport,clark2025molmo2, cho2025perceptionlm} demonstrate impressive performance on general benchmarks, their ``sense of time'' remains surprisingly brittle. As shown in Figure~\ref{fig:teaser}, even frontier models (e.g., GPT-5 and Gemini 3 Pro) struggle to distinguish atomic actions (e.g., \emph{shaking} vs. \emph{holding}) in videos as short as 10 seconds. Models often misjudge relative dynamics (e.g., \emph{accelerating} vs. \emph{decelerating}), and frequently fail to resolve linguistic temporal connectives (e.g., \emph{``as soon as''}). This discrepancy suggests that current benchmark scores~\cite{jang2017tgif, activitynet, li2024mvbench, zhou2025mlvu, fu2025video, hu2025video} severely overestimate genuine temporal understanding capabilities. 

The root issue lies in evaluation design. Existing benchmarks~\cite{activitynet, li2024mvbench,liu2024tempcompass,fu2025video} rarely \emph{isolate} temporal structure
as the sole discriminative factor. As a result, many models exploit ``static shortcuts''—correlating visual entities with answers without modeling time~\cite{lei2023revealing,krojer2025shortcut}. Furthermore, language priors~\cite{li2024naturalbench} allow models to guess answers based on textual plausibility. To truly \emph{diagnose} understanding, an evaluation must hold the visual content constant and vary only the temporal dynamics.

To address this, we introduce \textbf{TimeBlind}, a diagnostic benchmark for compositional spatio-temporal understanding (Table~\ref{tab:comparison}). Specifically, TimeBlind adopts a minimal-pairs design: each instance contains two videos with near-identical static visual content that differ \emph{solely} in their temporal structure. Furthermore, to eliminate language priors, we utilize complementary questions where the correct answer flips between the paired videos. By removing static and linguistic shortcuts, TimeBlind forces models to rely exclusively on temporal evidence. Unlike large-scale noisy benchmarks, TimeBlind prioritizes high-fidelity diagnostic precision. Similar to Winoground~\cite{thrush2022winoground}, each instance serves as a rigorous test for a specific cognitive primitive, where high-quality annotations are prioritized over scale.

Crucially, unlike most prior work~\cite{zhang2024vinoground, krojer2025shortcut}, TimeBlind evaluates logical composition, not just perception. Inspired by cognitive science, we extend the theory of \emph{Image Compositionality}~\cite{krishna2017visual, li2024genai} to the temporal domain, organizing the benchmark around a hierarchical taxonomy: (i) \emph{Events}, addressing what happened (e.g., changes in object attributes or action understanding); (ii) \emph{Event Attributes}, describing how events unfold (e.g., speed, magnitude of change); and (iii) \emph{Structural Event Logic}, examining how multiple events are composed (e.g., temporal topology, causality, and cross-event comparison). Within this framework, we encompass a diverse set of 11 fine-grained categories in real-world scenarios. For example, regarding \emph{Temporal Topology} in \emph{Structural Event Logic}, we incorporate Allen’s Interval Algebra~\cite{allen1983maintaining} by constructing videos that cover all 13 temporal event relations (e.g., \emph{overlaps}, \emph{meets}, \emph{equals}) and designing questions that distinguish specific topology differences within each pair, going beyond simple sequencing relations (e.g., \emph{before}, \emph{after}) used in prior work.

We evaluate over 20 state-of-the-art MLLMs, including leading proprietary models (e.g., Gemini 3 Pro~\cite{gemini}, GPT-5~\cite{openai2025gpt5systemcard}) as well as strong open-source models (e.g., Qwen3-VL~\cite{bai2025qwen3vltechnicalreport}, Molmo2~\cite{clark2025molmo2}, PLM~\cite{cho2025perceptionlm}), on a total of 600 curated instances (2,400 video-question pairs). We find that even the current top-performing model, Gemini 3 Pro, achieves only 48.2\% Instance Accuracy (I-Acc), which requires correctly distinguishing both videos in a pair, far below human performance (98.2\%). Furthermore, while advanced models like Gemini 3 Pro and GPT-5 demonstrate reasonable proficiency in recognizing isolated events (achieving 49.5\% and 58.3\% I-Acc, respectively), their performance degrades sharply on event attributes—such as Speed (slowly vs. rapidly), and Force (forcefully vs. gently)—dropping to 37.3\% and 32.4\% I-Acc. Moreover, our extensive ablation studies demonstrate that model performance remains poor even with increased input frames or test-time reasoning, as GPT-5 gains only 3.3\% in I-Acc, indicating that models remain “time-blind” and fragile in temporal reasoning. We believe that TimeBlind will serve as a vital diagnostic tool for evaluating and developing MLLMs capable of genuine temporal logic.

In summary, our main contributions are: 
\begin{itemize}[leftmargin=*,noitemsep,topsep=0pt,partopsep=0pt,parsep=0pt]
    \item \textbf{TimeBlind Benchmark.} A diagnostic, minimal-pairs benchmark that isolates temporal structure while minimizing static shortcuts and language priors during evaluation.
    \item \textbf{Taxonomy of Temporal Compositionality.} A cognitive hierarchy—\emph{Events}, \emph{Attributes}, and \emph{Structural Logic}—that guides systematic benchmark construction and evaluation.
    \item \textbf{Diagnostic Findings.} An evaluation of over 20 SOTA MLLMs revealing a significant gap between perceived and actual temporal reasoning capabilities across current models.
\end{itemize}

\section{Related Work}
\label{sec:related}

\begin{figure*}[t]
  \centering
  \includegraphics[width=1\linewidth]{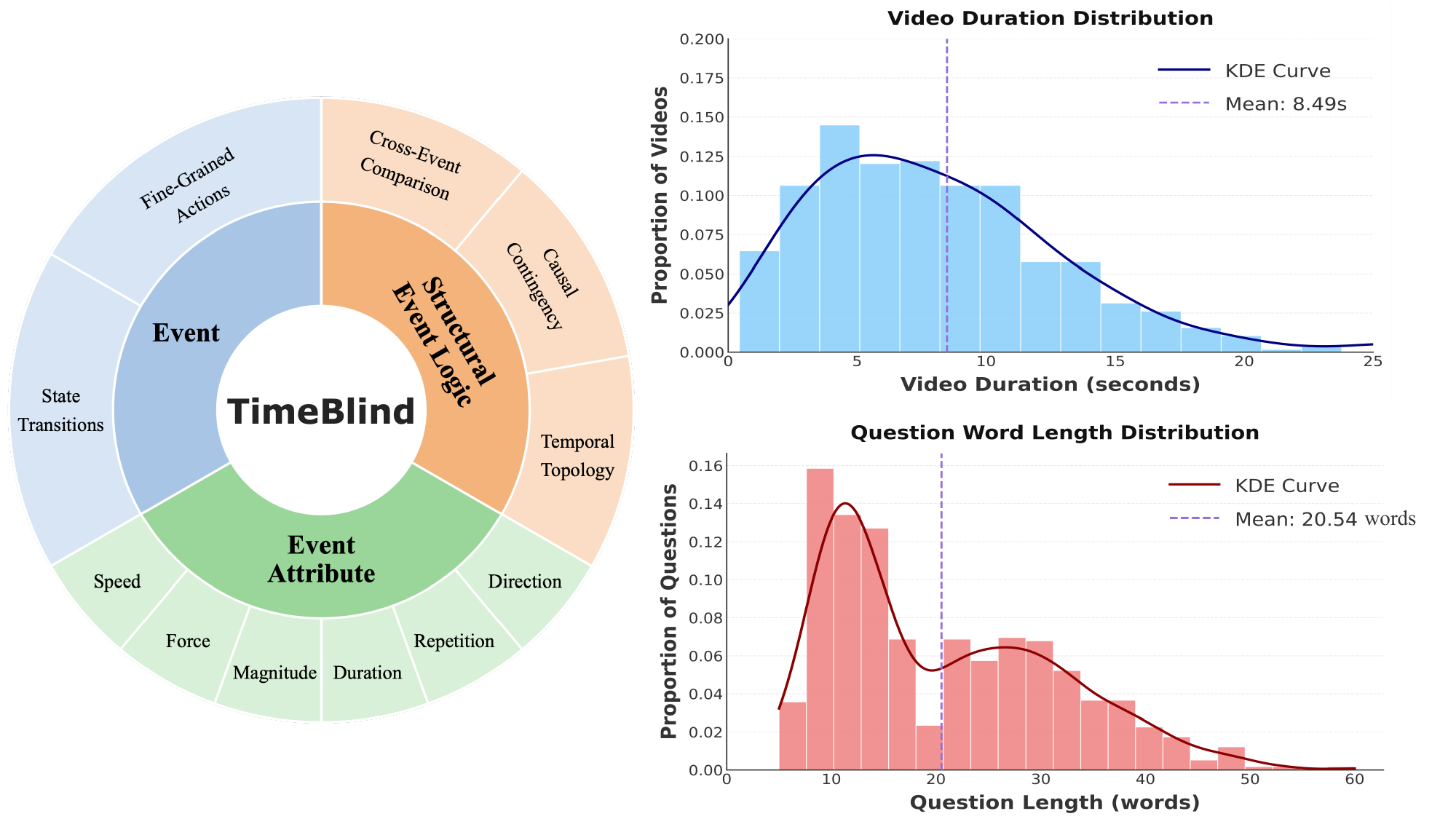}
  \caption{
  \textbf{TimeBlind Taxonomy and Statistics.} 
  \textbf{Left:} We structure the evaluation into 11 fine-grained spatio-temporal compositional categories spanning three high-level aspects: Atomic Events (\textit{what changes}), Parametric Event Attributes (\textit{how it changes}), and Structural Event Logic (\textit{how events compose}). 
  \textbf{Top Right:} Distribution of video lengths across the benchmark, showing that most videos fall within the 0--15 seconds. 
  \textbf{Bottom Right:} Distribution of question word counts, indicating that most questions are under 30 words.
  Overall, our benchmark features a structured taxonomy with diverse categories while maintaining short videos and concise questions.
}

  \label{fig:timeblind-taxonomy}
\end{figure*}

\noindent\textbf{VideoQA Benchmarks.} Early VideoQA datasets~\cite{activitynet, jang2017tgif, perception, xiao2021next, lei2018tvqa, xu2016msr, wu2024star} focus on simple scenarios with short clips and limited question types. Recent benchmarks address comprehensive evaluation~\cite{fu2025video, li2024mvbench, ma2025videoeval}, complex reasoning~\cite{nagrani2025minerva, cheng2025video, hu2025video, song2025video}, long-form understanding~\cite{mangalam2023egoschema, wang2025lvbench, wu2024longvideobench, song2024moviechat, zhou2025mlvu, yang2025egolife}, and domain-specific settings~\cite{pan2025basket, long2025adsqa, yi2025exact}. However, most do not isolate temporal structure as the sole discriminative factor, allowing models to exploit static shortcuts—relying on object co-occurrence or language priors without genuinely modeling temporal dynamics~\cite{lei2023revealing, krojer2025shortcut, li2024naturalbench}.

\noindent\textbf{Evaluating Spatio-Temporal Compositionality.} 
To assess the spatio-temporal understanding capabilities of MLLMs, several recent benchmarks have introduced diverse aspects of visual reasoning. 
In the image understanding domain, BLINK~\cite{fu2024blink} reformats and groups classic vision problems into perception-centric multiple-choice questions and carefully removes language priors.
In the video understanding domain, several works explicitly target temporal understanding by designing temporally challenging questions~\cite{johnson2017clevr, shangguan2024tomato, cai2024temporalbench, xue2025seeing}.
Beyond single-video evaluation, a growing line of work adopts paired video–question protocols to more precisely diagnose temporal reasoning.
\textsc{TempCompass}~\cite{liu2024tempcompass} constructs paired videos by systematically manipulating an original video, such as reversing playback or altering temporal speed, thereby isolating temporal understanding from ``static shortcuts.'' \textsc{VinoGround}~\cite{zhang2024vinoground} further strengthens this paradigm by requiring models to answer identical questions over paired videos, where the correct answer is determined solely by temporal differences, effectively removing language priors. 
Follow-up works like GLIMPSE~\cite{zhou2025glimpse} and MVP~\cite{krojer2025shortcut} scale this approach to test physical and visual-centric reasoning.
Different from prior works, \textsc{TimeBlind} achieves spatio-temporal compositionality through a carefully curated formal taxonomy.
Drawing inspiration from cognitive event perception~\cite{bach1986algebra} and extending the theory of image compositionality~\cite{krishna2017visual,thrush2022winoground,li2024naturalbench}, we decompose temporal reasoning into atomic \emph{Events}, parametric \emph{Event Attributes}, and \emph{Structural Logic}.

\section{The TimeBlind Benchmark}

\label{sec:benchmark}
\begin{figure*}[!t]
  \centering
  \includegraphics[width=\textwidth]{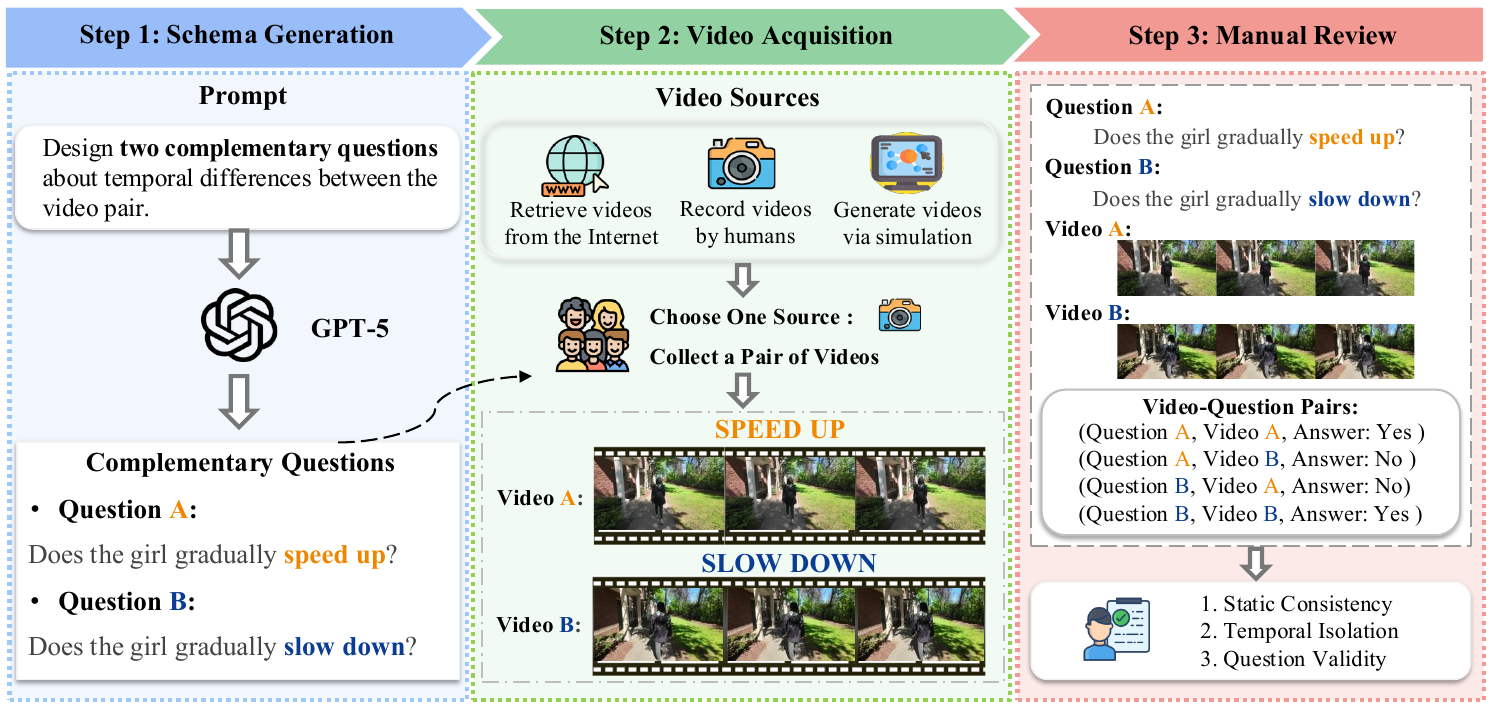}
\caption{\textbf{Overview of the TimeBlind data construction pipeline.} \textbf{Stage 1 (Schema Generation):} We prompt GPT-5 to generate paired complementary questions targeting temporal differences. \textbf{Stage 2 (Video Acquisition):} We collect one video pair that matches the generated schema from one of the following sources: (i) Retrieving videos from the internet, (ii) Recording videos with humans, or (iii) Generating videos via simulation (e.g., Unity). We then pair these videos with the questions to form a candidate TimeBlind instance. \textbf{Stage 3 (Manual Review):} Human annotators manually review each instance to ensure: (i) \textit{Static Consistency} (videos share identical static content), (ii) \textit{Temporal Minimality} (the pair differs only in the targeted temporal factor), and (iii) \textit{Question Validity} (QA pairs are clear and correct). }
  \label{fig:data_generation}
\end{figure*}

In this section, we describe the design of TimeBlind, a diagnostic benchmark for rigorously evaluating compositional spatio-temporal understanding. Our design is grounded in the following two principles: (i) a temporal minimal-pair protocol that minimizes visual shortcuts and language priors, and (ii) a cognition-inspired compositionality taxonomy that decomposes temporal reasoning into three primitives: \textit{Events}, \textit{Event Attributes}, and \textit{Structural Event Logic}.

\subsection{Task Formulation}
\label{sec:minpairs}
We structure each instance as $\mathcal{T} = (v_1, v_2, q_1, q_2)$, following the design below. This design creates a discriminative challenge where shortcut solutions are effectively ``canceled out,'' forcing the model to rely on temporal evidence.

\noindent\textbf{Temporal Minimal Pairs.}
The video pair $(v_1, v_2)$ constitutes a minimal pair: specifically, the videos share identical static content (e.g., objects, background) but differ along an isolated temporal axis defined in our taxonomy (Section~\ref{sec:taxonomy}). Consequently, a model cannot distinguish $v_1$ from $v_2$ without explicitly modeling the temporal features (e.g., \textit{opening} vs. \textit{closing}).

\noindent\textbf{Logical Complementarity.}
The questions $(q_1, q_2)$ are complementary: for any question, the ground-truth answer flips between the two videos (e.g., $Ans(v_1, q_j)\neq Ans(v_2, q_j)$ for $j\in\{1,2\}$). This design neutralizes language priors, preventing models from exploiting textual plausibility.

\noindent\textbf{Diagnostic Metrics.}
\label{sec:metrics}
Following rigorous diagnostic protocols~\cite{thrush2022winoground, li2024naturalbench}, we report a hierarchy of metrics to effectively mitigate shortcuts and more reliably assess the models' genuine temporal understanding:
\begin{itemize}[leftmargin=*,noitemsep,topsep=0pt,partopsep=0pt,parsep=0pt]
    \item \textbf{Accuracy (Acc):} Standard accuracy computed over all individual video–question trials.
    \item \textbf{Video Accuracy (V-Acc):} Measures visual consistency; correct only if the model answers \textit{both} questions correctly for a single video $v_j$. 
    \item \textbf{Question Accuracy (Q-Acc):} Measures textual consistency; correct only if the model answers question $q_i$ correctly for \emph{both} videos.
    \item \textbf{Instance Accuracy (I-Acc):} An instance is correct \textit{if and only if} the model solves all four trials. I-Acc is our primary proxy for spatio-temporal understanding, as it necessitates reliably distinguishing the temporal difference between the paired videos.
\end{itemize}

\subsection{A Hierarchy of Temporal Composition}
\label{sec:taxonomy}
To systematically evaluate temporal understanding, we carefully organize the TimeBlind benchmark around a compositional taxonomy mirroring bottom-up cognitive processing: from recognizing atomic changes (\textit{Events}), to measuring their intrinsic properties (\textit{Event Attributes}), and finally reasoning about their inter-dependencies (\textit{Structural Event Logic}), shown in Figure~\ref{fig:timeblind-taxonomy}.

\noindent\textbf{I. Events.} %
This foundational primitive tests the detection of atomic visual changes, demanding that models distinguish actual temporal evolution from static object presence in videos.
\begin{itemize}[leftmargin=*,noitemsep,topsep=0pt,partopsep=0pt,parsep=0pt]
    \item \textbf{Fine-Grained Actions:} Discriminating actions sharing visual 
context but differing in temporal dynamics (e.g., \emph{opening} vs. 
\emph{closing}) or semantic nuance (e.g., \emph{placing} vs. \emph{pushing}).
\item \textbf{State Transitions:} Tracking object property changes over time 
(shape, size, color, emotion)—e.g., liquid changing from clear to opaque.
\end{itemize}

\noindent\textbf{II. Event Attributes.} %
This primitive tests the perception of \emph{how} an event unfolds, requiring sensitivity to continuous or qualitative parameters rather than categorical recognition.
\begin{itemize}[leftmargin=*,noitemsep,topsep=0pt,partopsep=0pt,parsep=0pt]
    \item \textbf{Kinematics:} Properties governing motion in space-time, including \textit{Speed} (fast/slow), \textit{Direction} (forward/backward), \textit{Duration} (brief/long), and \textit{Repetition} (once/twice).
    \item \textbf{Dynamics \& Manner:} Properties reflecting physical forces or execution style, including \textit{Force} (gentle/forceful) and \textit{Magnitude} (slight/broad movement).
\end{itemize}

\noindent\textbf{III. Structural Event Logic.}
This level aims to test \emph{how} multiple events relate and compose into higher-order structures.
\begin{itemize}[leftmargin=*,noitemsep,topsep=0pt,partopsep=0pt,parsep=0pt]
    \item \textbf{Temporal Topology:} We adopt Allen’s Interval Algebra~\cite{allen1983maintaining} to comprehensively evaluate interval relations. Unlike prior work limited to simple sequencing (\emph{before/after}), we cover all 13 relations in our videos, such as  \emph{overlaps}, \emph{meets}, \emph{starts}, and \emph{during}.
    \item \textbf{Causal Contingency:} Identifying causal links between events \revise{(e.g., identifying which event \emph{caused} the effect)}.
    \item \textbf{Cross-Event Comparison:} Comparative reasoning across distinct events (e.g., ``Does the person hold the cup \emph{longer} than the book?'').
\end{itemize}

\subsection{Data Construction Pipeline}
\label{sec:construction}
We use a three-stage pipeline to collect high-quality TimeBlind data, as shown in Figure~\ref{fig:data_generation}.

\noindent\textbf{Step 1: Schema Generation.}
We prompt frontier LLMs (e.g., GPT-5) with the taxonomy definitions in Section \ref{sec:taxonomy} to generate structured specifications. Each output includes a pair of video descriptions (e.g., ``\textit{a girl walks in a park and gradually speeds up''} and ``\textit{a girl walks in a park and gradually slows down}'') that differ only in the target temporal factor (e.g., ``\textit{Speed}''), along with complementary questions (e.g., ``\textit{\newadd{Does the girl gradually speed up?}}'' and ``\textit{\newadd{Does the girl gradually slow down?}''}).

\noindent\textbf{Step 2: Video Acquisition.} We collect video pairs from three 
sources: (i) internet retrieval (extracting temporally distinct segments from 
the same source video), (ii) human recording, or (iii) simulation (e.g., Unity) 
for precise temporal control. We then pair the collected videos with the questions to form a candidate TimeBlind instance.

\noindent\textbf{Step 3: Rigorous Human Verification.}
\revise{We recruited 8 human annotators for quality control. Each instance was independently reviewed by two annotators and retained only if all constituent video-question pairs received unanimous approval. Annotators assessed three criteria: (i) \textit{Static Consistency}, ensuring that paired videos share identical static content; (ii) \textit{Temporal Minimality}, verifying that they differ only in the targeted temporal factor; and (iii) \textit{Question Validity}, confirming that the QA pairs are clear and correct. Overall, the inter-annotator agreement reached 94.3\%, with a Cohen's \(\kappa\) of 0.835. The entire review process took 62.2 hours in total. More detailed per-category agreement scores are reported in Appendix~\ref{app:quality_control}.}

\noindent\textbf{Statistics.}
TimeBlind comprises 2,400 curated video-question pairs, consisting of binary and two-choice multiple-choice questions. The video sources are distributed as follows: 24.0\% from Internet Retrieval, 57.7\% from Human Recording, and 18.3\% from Simulation. Regarding category distribution, we maintain an approximate 1:1:1 ratio across \textit{Event}, \textit{Event Attribute}, and \textit{Structural Event Logic}. Furthermore, within the \textit{Temporal Topology} sub-category, we ensure a balanced distribution across the 13 distinct Allen interval types.

\begin{table*}[t]
\centering
\caption{\textbf{Main Results on TimeBlind.} We use uniform sampling at 1 FPS and evaluate all models using default configurations. Metrics are reported following Section~\ref{sec:benchmark}, with I-Acc as the primary metric. The table is divided into \textit{Open-Source (grouped by size: $\leq 10$B and $\geq 30$B)}, and \textit{Closed-Source models}. Our results suggest that all models perform poorly on fine-grained temporal video understanding, with none of the methods achieving over 50\% I-Acc. Best results are shown in \textbf{bold}.}
\label{tab:overall}
\small
\setlength{\tabcolsep}{4pt}
\renewcommand{\arraystretch}{1.25}

\begin{tabularx}{\textwidth}{l >{\centering\arraybackslash}X c c c c}
\toprule
Model & Size & Q-Acc (\%) & V-Acc (\%) & Acc (\%) & I-Acc (\%) \\
\midrule
Random Chance & - & 25.0  & 25.0  & 50.0  & 6.3   \\
Human Evaluation & - & 98.8 & 98.9 & 99.3 & 98.2 \\
\midrule

\rowcolor{gray!15} \multicolumn{6}{c}{\textbf{\textit{Open-Source Models}}} \\

\midrule

\multicolumn{6}{l}{\textit{Model Size $\leq 10$B}} \\
\midrule
Keye-VL-1.5~\cite{team2025kwai}                 & 8B  & 17.5 & 26.5 & 55.6 & 6.6 \\
LLaVA-Video~\cite{zhang2024video}               & 7B  & 18.1 & 34.8 & 57.8 & 9.6 \\
InternVL 3.5~\cite{internvl3}                    & 8B  & 30.9 & 39.0 & 59.3 & 13.3 \\
PLM-8B~\cite{cho2025perceptionlm}                  & 8B  & 24.4 & 32.3 & 60.6 & 13.9 \\
F-16~\cite{li2025improving} & 7B & 23.5 & 39.7 & 60.5 & 14.5 \\
video-SALMONN 2+~\cite{tang2025video} & 7B & 26.3 & 41.7 & 61.7 & 16.2 \\
MiniCPM-V-4.5~\cite{yu2025minicpm} & 8B & 31.7 & 43.6 & 62.7 & 18.3 \\
VideoChat-Flash~\cite{li2024videochat} & 7B & 31.7 & 45.4 & 64.9 & 19.5 \\
Qwen3-VL~\cite{bai2025qwen3vltechnicalreport} & 8B & 34.5 & 43.6 & 64.8 & 19.7 \\
GLM-4.1V~\cite{hong2025glm} & 9B & 32.3 & 45.2 & 62.0 & 19.7 \\
Eagle2.5~\cite{chen2025eagle2.5} & 8B & 32.2 & 45.7 & 64.7 & 20.2 \\
Molmo2~\cite{clark2025molmo2} & 8B & \textbf{41.0} & \textbf{52.4} & \textbf{68.7} & \textbf{31.2} \\

\midrule

\multicolumn{6}{l}{\textit{Model Size $\geq 30$B}} \\
\midrule
InternVL 3.5~\cite{internvl3}                      & 38B  & 31.4 & 41.3 & 61.1 & 17.0 \\
LLaVA-Video~\cite{zhang2024video}                  & 72B  & 28.8 & 45.0 & 63.2 & 18.5 \\
Qwen3-VL~\cite{bai2025qwen3vltechnicalreport}      & 235B & \textbf{38.1} & \textbf{51.7} & \textbf{66.9} & \textbf{25.8} \\

\midrule

\rowcolor{gray!15} \multicolumn{6}{c}{\textbf{\textit{Closed-Source Models}}} \\
\midrule
Claude Sonnet 4.5~\cite{anthropic_claude_sonnet_4_5_2025}                                  & unknown   & 30.6 & 38.5 & 59.2 & 13.6 \\
Qwen3-VL Plus~\cite{bai2025qwen3vltechnicalreport}  & unknown   & 38.7 & 52.6 & 67.3 & 26.0 \\
GPT-5 mini~\cite{openai2025gpt5systemcard}                                          & unknown   & 42.3 & 53.7 & 68.3 & 30.0 \\
Gemini 2.5 Flash~\cite{google_gemini_2_5_thinking_2025}                                   & unknown   & 47.6 & 53.7 & 68.9 & 33.2 \\
Gemini 2.5 Pro~\cite{google_gemini_2_5_thinking_2025}                                   & unknown   & 56.0 & 62.6 & 74.2 & 43.5 \\
GPT-5~\cite{openai2025gpt5systemcard}               & unknown   & 58.5 & \textbf{67.2} & \textbf{77.3} & 46.3 \\
Gemini 3 Pro~\cite{gemini}                          & unknown   & \textbf{60.2} & 66.0 & 76.2 & \textbf{48.2} \\
\bottomrule
\end{tabularx}
\end{table*}

\section{Experimental Setup}
\label{sec:exp_setup}
\noindent\textbf{Baseline Models.} To thoroughly assess the challenges posed by the TimeBlind benchmark, we evaluate over 20 MLLMs, including both frontier proprietary models (e.g., GPT-5~\cite{openai2025gpt5systemcard}, Gemini 3 Pro~\cite{gemini}, Claude~\cite{anthropic_claude_sonnet_4_5_2025}, and Qwen3-VL Plus~\cite{bai2025qwen3vltechnicalreport}) and state-of-the-art open-source models (e.g., Qwen3-VL~\cite{bai2025qwen3vltechnicalreport}, InternVL~3.5~\cite{internvl3}, Molmo2~\cite{clark2025molmo2}, PLM~\cite{cho2025perceptionlm}, Keye-VL~\cite{team2025kwai}, MiniCPM-V-4.5~\cite{yu2025minicpm}, GLM-4.1V~\cite{hong2025glm}, and Eagle 2.5~\cite{chen2025eagle2.5}). 

\noindent\textbf{Evaluation Metrics.} We adopt the hierarchy of metrics defined in Section~\ref{sec:metrics}, including Standard Accuracy (Acc), Video Accuracy (V-Acc), Question Accuracy (Q-Acc), and our primary metric, Instance Accuracy (I-Acc). 

\noindent\textbf{Implementation Details.} \revise{We run most open-source model experiments on 4 NVIDIA H100 GPUs, while accessing proprietary models (e.g., GPT-5, Gemini 3 Pro) and large-scale open-source models (e.g., Qwen3-VL-235B) through their official APIs. Unless otherwise noted, we sample videos at 1 FPS (the default for most models) and evaluate in a zero-shot setting.}

\section{Experimental Results}

\label{sec:experiments}
\subsection{Main Results}
\label{sec:main_results}
In Table~\ref{tab:overall}, we present the overall model performance on our benchmark, revealing that all models perform poorly in fine-grained temporal video understanding. While many models achieve relatively high standard accuracy (Acc), they struggle to achieve good Instance Accuracy (I-Acc), which requires correctly distinguishing both videos in a pair. For instance, despite leading models like GPT-5 and Gemini 3 Pro achieving high Acc scores of 77.3\% and 76.2\% respectively, they only reach 46.3\% and 48.2\% on I-Acc. This discrepancy indicates that models remain weak in temporal understanding and suggests that high performance on Acc is often driven by shortcuts, rather than a true understanding of temporal dynamics. Another interesting finding is that Question Accuracy (Q-Acc) is consistently lower than Video Accuracy (V-Acc) across all evaluated models. This indicates that models are more prone to hallucinating answers based on textual patterns than misinterpreting visual cues. From the results, we also observe that among open-source models, Molmo2-8B—despite its smaller size—outperforms a range of sub-10B models and even surpasses the much larger Qwen3-VL-235B by 5.4\% on I-Acc, becoming the leading open-source model. This suggests that with effective model design, data curation, and training procedures, even smaller models can capture fine-grained temporal dynamics. Lastly, our results show a significant gap between open-source and proprietary models in understanding fine-grained temporal dynamics. Specifically, Molmo2-8B still lags substantially behind GPT-5 and Gemini 3 Pro, with gaps of 15.1\% and 17.0\% I-Acc, respectively.

\noindent\textbf{Human Evaluation.}  \revise{Four independent annotators evaluated the full TimeBlind under a balanced workload, with each seeing only one $(v, q)$ pair at a time and the four pairs from each instance distributed across different annotators (Table~\ref{tab:overall}). Humans reached 98.2\% I-Acc---50.0\% above Gemini 3 Pro---showing that TimeBlind's temporal dynamics are clear to humans yet challenging for MLLMs.}

\subsection{Category-Wise Diagnosis}
\label{sec:cat_results}
\revise{
We further analyze performance across the three high-level TimeBlind categories in Table~\ref{tab:category_short}. First, we observe that most models consistently perform better on discrete \emph{Events} than on \emph{Event Attributes} or \emph{Structural Event Logic}. For instance, GPT-5 reaches 58.3\% on \emph{Events} but drops to 32.4\% on \emph{Event Attributes}. Furthermore, performance is lowest in \emph{Event Attributes}, which probes continuous parameters such as speed, force, and magnitude. Even the strongest proprietary models GPT-5 and Gemini 3 Pro only achieve 32.4\% and 37.3\%, exposing a systematic deficiency in low-level, physics-related temporal understanding. Finally, a striking disparity emerges on \emph{Structural Event Logic} between proprietary and open-source models. While Gemini 3 Pro reaches 58.3\%, the \newaddd{strong} open-source model Qwen3-VL-235B only achieves 19.3\%. These results indicate that open-source MLLMs can detect isolated events but lack the logical abilities to reason about precise relationships between multiple events, particularly regarding causality.
}
\begin{table}[t]
\centering
\small
\setlength{\tabcolsep}{10pt}
\caption{\revise{\textbf{Category-Wise I-Acc (\%) on TimeBlind.} We report results across the three high-level categories: \textit{Event}, \textit{Attribute} (Event Attribute), and \textit{Logic} (Structural Event Logic). Models generally perform well on \textit{Events} but struggle with \textit{Attributes}, which require low-level physical understanding. Best results are \textbf{bolded}.}}
\label{tab:category_short}
\renewcommand{\arraystretch}{1}
\begin{tabular}{@{}lccc@{}}
\toprule
\textbf{Models} & \textbf{Event} & \textbf{Attribute} & \textbf{Logic} \\
\midrule
Random Chance & 6.3 & 6.3 & 6.3 \\
\midrule
\rowcolor{gray!10}
\multicolumn{4}{c}{\textit{Open-Source Models}} \\
\midrule
Qwen3-VL-235B     & 45.1 & 12.7 & \textbf{19.3} \\
Molmo2-8B         & \textbf{55.4} & \textbf{18.6} & 18.8 \\
\midrule
\rowcolor{gray!10}
\multicolumn{4}{c}{\textit{Closed-Source Models}} \\
\midrule
GPT-5        & \textbf{58.3} & 32.4 & 48.4 \\
Gemini 3 Pro & 49.5 & \textbf{37.3} & \textbf{58.3} \\
\bottomrule
\end{tabular}
\vspace{5pt}
\end{table}

\subsection{Shortcut Analysis}
To verify that TimeBlind requires genuine temporal understanding, we report three tests in Table~\ref{tab:temporal_tests}.

\newcolumntype{Y}{>{\centering\arraybackslash}X}
\begin{table}[t]
\centering
\setlength{\tabcolsep}{2pt}
\caption{
\textbf{Shortcut Analysis.}
We report I-Acc for three shortcut baselines using GPT-5 with 1 FPS sampling: Single Frame Bias (a question paired with only a randomly selected video frame), Language Only (a question without any visual input), and Shuffled Frames (a question with shuffled video frames). The results demonstrate that solving TimeBlind requires genuine temporal understanding.}
\label{tab:temporal_tests}
\small
\renewcommand{\arraystretch}{1.2}
\begin{tabularx}{\columnwidth}{l Y Y Y Y}
\toprule
\textbf{Setting} & \textbf{Acc} & \textbf{Q-Acc} & \textbf{V-Acc} & \textbf{I-Acc} \\
\midrule
Random Chance & 50.0 & 25.0 & 25.0 & 6.3 \\
\midrule
GPT-5~(\textit{baseline})  & 77.3 & 58.5 & 67.2 & 46.3 \\
\midrule
Single Frame   & 52.2 & 14.8 & 32.3 & 4.5 \\
Language Only   & 47.3 & 9.4  & 26.6 & 1.5 \\
\newadd{Shuffled Frames} & 49.6 & 10.7 & 23.2 & 3.0 \\
\bottomrule
\end{tabularx}
\end{table}

\begin{itemize}[leftmargin=*,noitemsep,topsep=0pt,partopsep=0pt,parsep=0pt]
\item \textbf{Single-Frame Bias.} This experiment evaluates whether TimeBlind requires reasoning over a sequence rather than exploiting information from a single static frame. In this setting, the model is provided with the question and only one randomly sampled frame. The results show that GPT-5 performs poorly (4.5\% I-Acc), indicating that TimeBlind requires sequential modeling.
\item \textbf{Language-Only Bias.} This experiment evaluates the influence of language priors within TimeBlind. In this setting, the model is provided with only the question without any visual information. As shown in Table~\ref{tab:temporal_tests}, GPT-5 achieves only 1.5\% I-Acc, demonstrating that visual information is essential for our benchmark setting. 
\item \textbf{\newadd{Shuffled Frames.}} Finally, we examine whether the benchmark can be solved by exploiting static visual cues that happen to correlate with temporal dynamics. In this setting, we randomly shuffle the order of the input frames sampled at 1 FPS to ensure that successfully completing the task requires an understanding of the temporal sequence rather than merely detecting specific objects. GPT-5 achieves only 3.0\% I-Acc in this setting, indicating that solving TimeBlind requires strict temporal understanding. 
\end{itemize}
Moreover, Acc scores across all three shortcut settings hover around random chance, suggesting that correctly answering even a single question in TimeBlind requires strict temporal understanding.

\subsection{Additional Analyses}
\label{sec:frames}
We conduct four ablations on model size, input FPS, inference-time reasoning, and input frames; more ablation studies are in Appendix~\ref{sec:more_ablation_study}.

\begin{table}[t!]
\centering
\small
\setlength{\tabcolsep}{10pt}
\caption{\revise{\textbf{Effect of Model Size.} We report the I-Acc of Qwen3-VL across three model sizes under the same video input settings. Scaling from 8B to 235B improves I-Acc to 25.8\% (by 6.1\%), still trailing human performance by 72.4\%, indicating that TimeBlind remains highly challenging even for large-scale models.}}
\label{tab:model_size_qwen}
\begin{tabular}{@{}lccc@{}}
\toprule
\textbf{Model Size} & 8B & 30B & 235B \\
\midrule
\textbf{I-Acc} & 19.7 & 22.5 & \textbf{25.8} \\
\bottomrule
\end{tabular}
\vspace{6pt}
\end{table}

\noindent\textbf{Effect of Model Size.} 
\newadd{We study the effect of model size under the same input and evaluation settings. Table~\ref{tab:model_size_qwen} shows that scaling up model size alone is not sufficient: Qwen3-VL (evaluated at 1 FPS) improves by 6.1\% in I-Acc when scaled from 8B to 235B, still trailing human performance by 72.4\%. Simply increasing model size does not yield robust spatio-temporal understanding.}

\begin{table}[tb!]
\caption{\revise{\textbf{Effect of Input FPS.} We report I-Acc across different input frame rates. The best score—Gemini 3 Pro at 5 FPS (56.2\%)—still trails human performance by 42.0\%. All models peak at 5 FPS and degrade at 10 FPS, indicating that higher FPS does not consistently improve temporal understanding.}}
\renewcommand{\arraystretch}{1.1}
\centering
\small
\setlength{\tabcolsep}{8pt}
\begin{tabular}{lccc}
\toprule
\textbf{Model} & \textbf{1 FPS} & \textbf{5 FPS} & \textbf{10 FPS} \\
\midrule
Human & 98.2 & 98.2 & 98.2 \\
\midrule
Qwen3-VL-235B & 25.8 & 35.0 & 26.5 \\
Molmo2-8B & 31.2 & 31.7 & 29.2 \\
GPT-5 & 46.3 & 51.0 & 49.3 \\
Gemini 3 Pro & 48.2 & 56.2 & 49.7 \\
\bottomrule
\end{tabular}
\label{tab:fps}
\end{table}

\noindent\textbf{Effect of FPS.}
We report I-Acc across a range of different input frame rates. Increasing FPS from 1 to 5 brings substantial gains, but the best score—Gemini 3 Pro at 5 FPS (56.2\%)—still trails human performance by 42.0\%. All models peak at 5 FPS and degrade at 10 FPS, showing that denser temporal sampling does not consistently improve performance under our evaluation setup.

\begin{table}[t!]
    \renewcommand{\arraystretch}{1.1}
    \small
    \centering
        \caption{\revise{\textbf{Inference-Time Reasoning.} We compare standard Qwen3-VL and GPT-5 models against their reasoning-enabled \emph{Thinking} counterparts across various settings, using the I-Acc metric at 1 FPS sampling. Results show that inference-time reasoning depth improves temporal understanding, but is far from sufficient to solve TimeBlind.}}
    \resizebox{\columnwidth}{!}{ 
    \begin{tabular}{l l l r}
        \toprule
        \textbf{Model} & \textbf{Mode} & \textbf{I-Acc (\%)} & \textbf{$\Delta$} \\
        \midrule
        \textit{Qwen3-VL-8B}  & Standard & 19.4 & - \\
        \rowcolor{gray!15} 
                      & Thinking & 27.8 & +8.4 \\
        \midrule
        \textit{Qwen3-VL-235B} & Standard & 25.6 & - \\
        \rowcolor{gray!15} 
                      & Thinking & 36.1 & \textbf{+10.5} \\
        \midrule
        \textit{GPT-5}        & Standard & 46.7 & - \\
        \rowcolor{gray!15} 
                      & Thinking & \textbf{50.0} & +3.3 \\
        \bottomrule
    \end{tabular}
    }
    \label{tab:thinking_ablation}
    \vspace{6pt}
\end{table}

\noindent\textbf{Test-Time Scaling (Reasoning).} \newadd{We evaluate the effect of enhanced reasoning on model performance, with results shown in Table~\ref{tab:thinking_ablation}.} Due to computational cost, we evaluate on a random 30\% subset of the data. For Qwen3-VL, \emph{Thinking} variants outperform \emph{Instruct} models, with the 235B model gaining 10.5\% but still achieving only 36.1\% I-Acc. For GPT-5, \emph{Thinking} yields modest gains, achieving 50.0\% I-Acc—still far below human performance (98.2\%). While reasoning helps, it remains insufficient to solve TimeBlind.

\noindent\textbf{Effect of Input Frames.} We evaluate several open-source and proprietary models under uniform sampling with different frame counts. Table~\ref{tab:analysis_frames_improved} shows that models' spatio-temporal understanding remains weak even as the frame count increases from 8 to 32: InternVL 3.5 (38B) reaches 25.2\% I-Acc at 32 frames, far below human performance (98.2\%); GPT-5 gains 1.6\% at 16 frames over 8 frames, and its performance at 32 frames (48.3\%) even falls below that at 8 frames (49.2\%).

\begin{table}[bt!]
\centering
\caption{\textbf{Effect of Input Frames.} We report the I-Acc performance of several open-source and proprietary models across various input frame counts under uniform sampling. These results show that additional frames can benefit some models, but their spatio-temporal understanding remains weak. For example, InternVL 3.5 (38B) reaches 25.2\% I-Acc at 32 frames, far below human performance (98.2\%). Best results within each model grouping are highlighted in \textbf{bold}.}
\label{tab:analysis_frames_improved}
\small
\setlength{\tabcolsep}{4pt}
\renewcommand{\arraystretch}{1.2}
\newcolumntype{Y}{>{\centering\arraybackslash}X}
\begin{tabularx}{\columnwidth}{c Y Y Y Y}
\toprule
\textsc{Input} & \multicolumn{4}{c}{\textsc{Results}} \\
\cmidrule(lr){1-1} \cmidrule(lr){2-5}
\textbf{Frames} & \textbf{Q-Acc} & \textbf{V-Acc} & \textbf{Acc} & \textbf{I-Acc} \\
\midrule
\rowcolor{gray!15} \multicolumn{5}{c}{\textit{InternVL 3.5 (38B)}} \\
8  & 35.8 & 42.8 & 63.2 & 20.3 \\
16 & 38.9 & 45.4 & 64.5 & 24.3 \\
32 & \textbf{39.4} & \textbf{46.4} & \textbf{65.9} & \textbf{25.2} \\
\midrule
\rowcolor{gray!15} \multicolumn{5}{c}{\textit{LLaVA-Video (72B)}} \\
8  & 29.7 & 46.2 & 63.5 & 19.8 \\
16 & 32.4 & 48.4 & 65.0 & 23.0 \\
32 & \textbf{32.8} & \textbf{48.7} & \textbf{65.2} & \textbf{23.3} \\
\midrule
\rowcolor{gray!15} \multicolumn{5}{c}{\textit{GPT-5}}\\
8  & 60.9 & 69.2 & 79.0 & 49.2 \\
16 & \textbf{62.7} & \textbf{69.9} & \textbf{79.5} & \textbf{50.8} \\
32 & 61.1 & 68.2 & 78.7 & 48.3 \\
\bottomrule
\end{tabularx}
\end{table}

\section{Conclusion}
We introduce TimeBlind, a diagnostic benchmark for compositional spatio-temporal reasoning in MLLMs. Through a cognitive taxonomy—spanning atomic \textit{events}, \textit{event attributes}, and \textit{structural event logic}—and a strict minimal-pair design, TimeBlind isolates temporal understanding from static and linguistic shortcuts. Our evaluation reveals that despite rapid progress on static vision-language tasks, state-of-the-art models remain largely ``time-blind'', trailing human performance by 50.0\% and struggling with fine-grained event attributes and temporal logic. We hope TimeBlind will support the development of MLLMs with genuine temporal understanding.

\section*{Limitations}
\revise{
While TimeBlind offers a rigorous diagnostic for compositional spatio-temporal understanding, we acknowledge several limitations. First, TimeBlind focuses on short-form videos (average 8.49 seconds), which isolates fine-grained temporal dynamics but leaves long-horizon temporal reasoning over minutes or hours outside its scope. Second, all questions are written in English, leaving the assessment of multilingual temporal understanding as an open direction. Third, we adopt binary and two-choice multiple-choice formats to enable unambiguous scoring under our minimal-pair protocol; evaluating temporal understanding through open-ended generation remains a complementary direction. Finally, certain \emph{Event Attribute} sub-categories (e.g., Force, Magnitude) inherently involve qualitative judgments whose boundaries can be subjective; while our annotation pipeline maintains substantial inter-annotator agreement ($\kappa = 0.84$ for \emph{Event Attributes}), residual ambiguity is intrinsic to these concepts. We view these as natural directions for extending TimeBlind, rather than as challenges to its diagnostic conclusions.
}

\section*{Acknowledgments} This work was supported by the Laboratory for Analytic Sciences via NC State University, ONR Award N00014-23-1-2356, National Institutes of Health Award 1R01HD111074-01, a Sony Focused Research award and NSF CAREER Award. We thank Zhiqiu Lin, Kewen Wu, and Deva Ramanan for their invaluable discussions during the development of this work.

\newpage
\bibliography{main.bib}

\clearpage
\newpage

\appendix
\section{Appendix}
\label{sec:appendix}

This document supplements the main paper with additional details on the TimeBlind benchmark, extended analyses, and representative failure cases. The contents are organized as follows:
\begin{itemize}[leftmargin=*,itemsep=2pt,topsep=0pt,partopsep=0pt,parsep=0pt]
\item {\bf Benchmark Details} (Sections~\ref{sec:app_timeblind}--\ref{sec:schema_robustness}): data collection, prompts for data generation, quality control, benchmark statistics, taxonomy details, and schema generator robustness.
\item {\bf Additional Analyses} (Sections~\ref{sec:cat_results_more} and~\ref{sec:more_ablation_study}): category-wise diagnosis, and ablations on video similarity and video sources.
\item {\bf Failure Cases} (Section~\ref{app_sec:failure_cases}): representative failure examples of state-of-the-art models.
\item {\bf Impact Statement} (Section~\ref{sec:impact}): a discussion of the broader impact of TimeBlind.
\item {\bf License and Intended Use} (Section~\ref{sec:license}): license terms, intended use, and use of existing artifacts.
\end{itemize}

\subsection{Data Collection Processes}
\label{sec:app_timeblind}

\noindent\textbf{Details on human-recorded videos.} We recruited 10 participants and provided each with generated caption pairs and clear instructions: (1) record two videos matching the given captions; (2) maintain static consistency (same people, props, scene, camera angle); (3) ensure temporal minimality (videos differ only in the targeted temporal factor); and (4) meet quality standards (clear footage, similar duration within pairs). Videos are not limited to single-person scenes—some include multi-person interactions across settings such as parks, streets, restaurants, and offices. 

\noindent\textbf{Details on internet video collection.} Annotators searched YouTube using LLM-generated captions or relevant keywords. They manually identified clips matching caption pairs. Where possible, both clips were sourced from the same source video to maximize static consistency.

\noindent\textbf{Details on generated video collection.} We used tools such as Unity to create videos for each caption pair. During generation, we ensured static visual consistency within each video pair, introduced differences only in the targeted temporal aspect, and kept the two videos in each pair nearly the same in duration.

\subsection{Quality Control Process}\label{app:quality_control}We used 8 annotators for quality review (separate from the 10 video recorders and 4 human evaluators). Each instance was independently reviewed by 2 annotators, and video-question pairs in each instance were accepted only upon unanimous agreement. Instances were kept only if all constituent pairs passed. The inter-annotator agreement was 94.3\%, and Cohen's kappa was 0.835. The total review time was 62.2 hours (\(\sim\)2-3 min per instance per annotator). We also report the per-category annotator agreement scores in Table~\ref{tab:human_agreement}; all three categories achieve substantial agreement ($\kappa > 0.8$). Event Attributes show the lowest agreement, consistent with the inherently more subjective nature of judging continuous properties such as speed and force, but still well above conventional thresholds for high-quality annotation. The final benchmark was then evaluated by 4 separate, independent annotators who achieved 98.2\% I-Acc, confirming that it is high-quality and largely unambiguous.

\begin{table}[h]
\small
\centering
\caption{Human annotation agreement across categories.}
\label{tab:human_agreement}
\begin{tabular}{lcc}
\toprule
\textbf{Category} & \textbf{Agreement} & \textbf{Cohen's $\kappa$} \\
\midrule
Events                 & 95.6\% & 0.88 \\
Event Attributes       & 93.4\% & 0.84 \\
Structural Event Logic & 93.7\% & 0.81 \\
\bottomrule
\end{tabular}
\end{table}

\subsection{Benchmark Statistics}
\noindent\textbf{Data Statistics.}
\label{app_sec:benchmark_statistics} TimeBlind includes both two-choice multiple-choice and binary questions in equal numbers. It is a short-video benchmark, with an average video duration of 8.49 seconds. We computed CLIP-based video similarity for all pairs by sampling frames at 1 FPS, extracting per-frame embeddings, and average-pooling within each video. The mean pair similarity is 0.971 (median 0.979), and 88.3\% of instances have similarity above 0.95, confirming strong static consistency. 
\begin{table}[h!]
    \centering
    \small
    \caption{\textbf{Statistics of the TimeBlind benchmark.}}
    \label{tab:statistics}
    \begin{tabular}{lc}
        \toprule
        \textbf{Statistic} & \textbf{Value} \\
        \midrule
        Binary Question & 1,200 \\
        Two-Choice Multiple-Choice Question & 1,200 \\
        \midrule
        Avg. Video Length (sec) & 8.49 \\
        Avg. Question Length (words) & 20.54 \\
        \bottomrule
    \end{tabular}
\end{table}

\noindent\textbf{Statistical significance.} We report statistical significance tests below. Wilson 95\% confidence intervals on I-Acc ($n=600$): (1)~Random chance: 6.3\% [4.6\%, 8.5\%], (2)~Best open-source (Molmo2-8B): 31.2\% [27.6\%, 35.0\%], (3)~Best overall (Gemini 3 Pro): 48.2\% [44.2\%, 52.2\%], (4)~Human: 98.2\% [96.8\%, 99.0\%].

All intervals are non-overlapping, confirming that performance differences across tiers are statistically significant. Since all models are evaluated on the same 600 instances, we use the paired McNemar's test for pairwise comparisons. The gap between Gemini 3 Pro (48.2\%) and human performance (98.2\%) is highly significant ($p = 6.69 \times 10^{-81}$). The gap between Gemini 3 Pro and the best open-source model, Molmo2-8B (31.2\%), is likewise significant ($p = 1.67 \times 10^{-12}$). A one-sample exact binomial test of Gemini 3 Pro against random chance (6.3\%) confirms performance significantly exceeds chance ($p = 1.91 \times 10^{-178}$). Even the small 1.9\% gap between GPT-5 (46.3\%) and Gemini 3 Pro (48.2\%) is correctly found to be nonsignificant (McNemar's $p = 0.441$), demonstrating that our sample size has appropriate discriminative power: it captures meaningful gaps while not over-claiming marginal ones.

\subsection{Prompt Details.} In this section, we present the full prompt template used in our Human--AI collaborative pipeline (see Section~\ref{sec:construction}).
The system prompt for TimeBlind consists of three parts: (1) a \textbf{General Prompt} that defines the role and task, (2) \textbf{Detailed Requirements} that specify the core generation rules and logical constraints, and (3) an \textbf{Output Format} that standardizes the response structure. The prompt includes strict structural constraints to ensure high-quality minimal pairs. When using this prompt, we provide a single category as the ``Input'' each time.

\begin{tcolorbox}[
    enhanced,
    title=\textbf{Part 1: Role, Context, \& Task Definition},
    label={box:part1},
    colback=gray!2,
    colframe=blue!35!black,
    colbacktitle=blue!8,
    coltitle=black,
    boxrule=0.5pt,
    arc=2pt,
    left=8pt,
    right=8pt,
    top=10pt,
    bottom=8pt,
    fonttitle=\bfseries,
    fontupper=\small,
    attach boxed title to top left={xshift=3mm,yshift=-2mm},
    boxed title style={
        colback=blue!8,
        colframe=blue!35!black,
        boxrule=0.5pt,
        arc=2pt,
        left=5pt,
        right=5pt,
        top=2pt,
        bottom=2pt
    },
    before skip=0.6em,
    after skip=0.6em
]

\noindent{\color{blue!45!black}\textbf{Role.}}
You are an expert in dataset curation and benchmark design for Multimodal Large Language Models (MLLMs).

\vspace{0.5em}
\noindent{\color{blue!45!black}\textbf{Context.}}
We are constructing a benchmark called TimeBlind to evaluate the temporal perception capabilities of MLLMs. Each TimeBlind instance consists of two videos $(v_1, v_2)$ and two questions $(q_1, q_2)$.

\begin{itemize}[leftmargin=1.3em, itemsep=2pt, topsep=3pt, parsep=0pt]
    \item The videos are visually similar, sharing the same objects and background, but differ in temporal dynamics.
    \item The questions are designed to distinguish these specific temporal features.
    \item During testing, the model receives only \textbf{one video} and \textbf{one question} at a time.
\end{itemize}

\vspace{0.5em}
\noindent{\color{blue!45!black}\textbf{Task.}}
Given a specific Temporal Dynamic Category, generate a valid data instance consisting of one pair of videos and one pair of questions that satisfies the requirements below.

\end{tcolorbox}

\begin{tcolorbox}[
    enhanced,
    title=\textbf{Part 2: Requirements \& Constraints},
    label={box:part2},
    colback=gray!2,
    colframe=blue!35!black,
    colbacktitle=blue!8,
    coltitle=black,
    boxrule=0.5pt,
    arc=2pt,
    left=8pt,
    right=8pt,
    top=10pt,
    bottom=8pt,
    fonttitle=\bfseries,
    fontupper=\small,
    attach boxed title to top left={xshift=3mm,yshift=-2mm},
    boxed title style={
        colback=blue!8,
        colframe=blue!35!black,
        boxrule=0.5pt,
        arc=2pt,
        left=5pt,
        right=5pt,
        top=2pt,
        bottom=2pt
    },
    before skip=0.6em,
    after skip=0.6em
]

\noindent{\color{blue!45!black}\textbf{Requirements.}}

\vspace{0.4em}
\noindent\textit{1. Video Generation $(v_1, v_2)$.}
\begin{itemize}[leftmargin=1.3em, itemsep=2pt, topsep=3pt, parsep=0pt]
    \item \textbf{Format:} Provide detailed text descriptions for both videos.
    \item \textbf{Feasibility:} The scenes should be easy to collect via web search, self-recording, or synthetic generation.
    \item \textbf{Visual Consistency:} $v_1$ and $v_2$ must share the same objects, background, and camera angle. They should look nearly identical in a static frame, with differences only in temporal dynamics.
\end{itemize}

\vspace{0.5em}
\noindent\textit{2. Question Generation $(q_1, q_2)$.}
\begin{itemize}[leftmargin=1.3em, itemsep=2pt, topsep=3pt, parsep=0pt]
    \item \textbf{Type:} Use either two-choice multiple-choice questions or binary questions (Yes/No).
    \item \textbf{Relevance:} Both questions must relate to the given ``Temporal Dynamic Category.''
    \item \textbf{Temporal Dependency:} The questions must require temporal understanding and cannot be answered from a single static frame.
\end{itemize}

\vspace{0.7em}
\noindent{\color{blue!45!black}\textbf{Logical Constraints.}}

The answers $A$ must satisfy the following constraints to ensure that the model is truly perceiving time.

\vspace{0.3em}
\noindent\textit{Constraint A: Cross-Video Difference.}  
The answer to each question must change across videos:
\[
\begin{aligned}
A(q_1, v_1) &\neq A(q_1, v_2), \\
A(q_2, v_1) &\neq A(q_2, v_2).
\end{aligned}
\]

\noindent\textit{Constraint B: Intra-Video Difference.}  
The two questions must have different answers within the same video:
\[
\begin{aligned}
A(q_1, v_1) &\neq A(q_2, v_1), \\
A(q_1, v_2) &\neq A(q_2, v_2).
\end{aligned}
\]

\end{tcolorbox}

\begin{tcolorbox}[
    enhanced,
    title=\textbf{Part 3: Output Specifications},
    label={box:part3},
    colback=gray!2,
    colframe=blue!35!black,
    colbacktitle=blue!8,
    coltitle=black,
    boxrule=0.5pt,
    arc=2pt,
    left=8pt,
    right=8pt,
    top=10pt,
    bottom=8pt,
    fonttitle=\bfseries,
    fontupper=\small,
    attach boxed title to top left={xshift=3mm,yshift=-2mm},
    boxed title style={
        colback=blue!8,
        colframe=blue!35!black,
        boxrule=0.5pt,
        arc=2pt,
        left=5pt,
        right=5pt,
        top=2pt,
        bottom=2pt
    },
    before skip=0.6em,
    after skip=0.6em
]

\noindent{\color{blue!45!black}\textbf{Output Format.}}

Present the result using the following structured format.

\vspace{0.5em}
\noindent\textbf{Temporal Category:} [Input Category]

\vspace{0.6em}
\noindent\textbf{Video Descriptions}
\begin{itemize}[leftmargin=1.3em, itemsep=2pt, topsep=3pt, parsep=0pt]
    \item \textbf{$v_1$:} [Description]
    \item \textbf{$v_2$:} [Description]
    \item \textbf{Collection Source:} [Synthetic / Self-shot / Web]
\end{itemize}

\vspace{0.5em}
\noindent\textbf{Questions \& Answers}
\begin{itemize}[leftmargin=1.3em, itemsep=3pt, topsep=3pt, parsep=0pt]
    \item \textbf{$q_1$:} [Question Text] \\
    \textit{Answer in $v_1$:} [Answer]; \quad
    \textit{Answer in $v_2$:} [Answer]

    \item \textbf{$q_2$:} [Question Text] \\
    \textit{Answer in $v_1$:} [Answer]; \quad
    \textit{Answer in $v_2$:} [Answer]
\end{itemize}

\vspace{0.5em}
\noindent\textbf{Reason}
\begin{itemize}[leftmargin=1.3em, itemsep=2pt, topsep=3pt, parsep=0pt]
    \item \textbf{Why a single frame is insufficient:} [Brief explanation]
\end{itemize}

\end{tcolorbox}

\subsection{Taxonomy}
\label{sec:app_taxonomy}
In this section, we introduce the detailed taxonomy of TimeBlind as summarized in Table~\ref{tab:video_skills}. 

\newcolumntype{V}[1]{>{\raggedright\arraybackslash}m{#1}} %
\newcolumntype{E}[1]{>{\raggedright\arraybackslash}p{#1}} %

\setlist[itemize]{leftmargin=*, nosep, topsep=0pt, partopsep=0pt}
\newcommand{\exitems}[1]{\begin{itemize}#1\end{itemize}}

\setlength{\tabcolsep}{8pt}
\renewcommand{\arraystretch}{1.0}

\begin{table*}[h!]
\centering
\caption{\textbf{Hierarchical taxonomy of TimeBlind for video understanding.}
The taxonomy is structured into three levels—Event, Event Attribute, and Structural Event Logic. For each level, we provide its conceptual definition, associated subcategories, and representative paired video descriptions.}

\scalebox{0.9}{

\begin{NiceTabular}{V{0.15\linewidth} V{0.19\linewidth} E{0.23\linewidth} E{0.36\linewidth}}
\toprule[1.2pt]
\textbf{Taxonomy} & \textbf{Definition} & \textbf{Sub-Categories} & \textbf{Example: Paired Video Descriptions} \\ \midrule

\Block[l]{2-1}{{\fontfamily{cmtt}\selectfont Event}}
&
\Block[l]{2-1}{The atomic units of temporal understanding, encompassing fine-grained action phases and diverse state transitions.}
& \textbf{Fine-grained Action} \newline {\footnotesize Subtle differences in action phase that are difficult to resolve from a single frame.}
& \exitems{
    \item {\it The man {\bf opens} the door.}
    \item {\it The man {\bf closes} the door.}
  } \\ \cmidrule{3-4}
& & \textbf{State Transitions} \newline {\footnotesize Tracking attribute changes over time (e.g., color, size, shape).}
& \exitems{
    \item {\it The apple turns {\bf from red to yellow}.}
    \item {\it The apple turns {\bf from yellow to red}.}
  } \\ \midrule

\Block[l]{6-1}{{\fontfamily{cmtt}\selectfont Event Attribute}}
&
\Block[l]{6-1}{The properties that characterize how an event unfolds over time.}
& \textbf{Speed} \newline {\footnotesize The rate at which an action is performed or an object moves over time.}
& \exitems{
    \item {\it The man {\bf quickly} opens the door.}
    \item {\it The man {\bf slowly} opens the door.}
  } \\ \cmidrule{3-4}
& & \textbf{Force} \newline {\footnotesize The intensity of an action as reflected in visible physical effects (e.g., acceleration, impact strength, or deformation).}
& \exitems{
    \item {\it The man {\bf gently} opens the door.}
    \item {\it The man {\bf forcefully} opens the door.}
  } \\ \cmidrule{3-4}
& & \textbf{Magnitude} \newline {\footnotesize The spatial extent or scale of motion or deformation.}
& \exitems{
    \item {\it The person waves {\bf slightly}.}
    \item {\it The person waves {\bf broadly}.}
  } \\ \cmidrule{3-4}
& & \textbf{Duration} \newline {\footnotesize The length of time an event lasts.}
& \exitems{
    \item {\it The person holds the cup for a {\bf long time}.}
    \item {\it The person holds the cup {\bf briefly}.}
  } \\ \cmidrule{3-4}
& & \textbf{Direction} \newline {\footnotesize The trajectory or spatial direction of motion.}
& \exitems{
    \item {\it The person moves {\bf toward the right side of the frame}.}
    \item {\it The person moves {\bf toward the left side of the frame}.}
  } \\ \cmidrule{3-4}
& & \textbf{Repetition} \newline {\footnotesize The number of times an event occurs.}
& \exitems{
    \item {\it The person picks up the cup {\bf three times}.}
    \item {\it The person picks up the cup {\bf twice}.}
  } \\ \midrule

\Block[l]{3-1}{{\fontfamily{cmtt}\selectfont Structural Event Logic}}
&
\Block[l]{3-1}{The higher-level rules governing how events relate to and compose with one another.}
& \textbf{Temporal Topology} \newline {\footnotesize Interval relations between event intervals (Allen’s interval algebra), such as before, meet, overlap, start, during, finish, and equal.}
& \exitems{
    \item {\it The person picks up the book and cup {\bf at the same time}.}
    \item {\it The person picks up the book and subsequently picks up the cup {\bf after a short delay}.}
  } \\ \cmidrule{3-4}
& & \textbf{Causal Contingency} \newline {\footnotesize The causal dependencies between events.}
& \exitems{
    \item {\it The {\bf first press} causes the light to turn off.}
    \item {\it The {\bf second press} causes the light to turn off.}
  } \\ \cmidrule{3-4}
& & \textbf{Cross-Event Comparison} \newline {\footnotesize The comparative reasoning over attributes of different events within the same video.}
& \exitems{
    \item {\it The person holds the cup {\bf longer than} the book.}
    \item {\it The person holds the cup for {\bf less time than} the book.}
  } \\
\bottomrule[1.2pt]
\end{NiceTabular}
}
\label{tab:video_skills}
\end{table*}

\subsection{Robustness of Schema Generator}
\label{sec:schema_robustness}

In this section, we examine the robustness of the schema generator used in our data construction pipeline (Section~\ref{sec:construction}). First, schema generation was guided by human-written in-context examples, reducing GPT-5-specific stylistic artifacts. Furthermore, to verify that our benchmark does not depend on GPT-5-generated schemas, we collected a new set of 50 instances with schemas written by four human annotators, and compared model performance on this set against the GPT-5-generated schema set. As shown in Table~\ref{tab:schema_robustness}, the model rankings are consistent across the two sets, and GPT-5 exhibits no self-preference on data generated from its own schemas, achieving similar I-Acc on both sets (46.3\% vs. 48.0\%). These results indicate that our findings are robust to the choice of schema source.

\begin{table}[t]
\centering
\caption{\textbf{Comparison of model performance on GPT-5-generated and human-written schemas.} The table reports results in I-Acc (\%). Model rankings remain consistent across the two schema sources, and GPT-5 shows no self-preference on data generated from its own schemas.}
\label{tab:schema_robustness}
\resizebox{\columnwidth}{!}{%
\begin{tabular}{lcc}
\toprule
\textbf{Model} & \textbf{GPT-5 Schema} & \textbf{Human-Written Schema} \\
\midrule
Qwen3-VL-235B & 25.8 & 22.0 \\
GPT-5         & 46.3 & 48.0 \\
Gemini 3 Pro  & \textbf{48.2} & \textbf{52.0} \\
\bottomrule
\end{tabular}%
}
\vspace{-5pt}
\end{table}

\begin{table*}[t!]
\centering
\small
\setlength{\tabcolsep}{4pt}
\renewcommand{\arraystretch}{1.1}

\caption{\textbf{Category-Wise I-Acc (\%) on TimeBlind.} This table reports I-Acc for advanced models across 11 fine-grained temporal understanding tasks to pinpoint specific cognitive deficits. Due to space constraints, we use the following abbreviations for temporal categories: \textit{\textbf{FG Action}}: Fine-Grained Action, \textit{\textbf{State Trans}}: State Transitions, \textit{\textbf{Mag}}: Magnitude, \textit{\textbf{Dir}}: Direction, \textit{\textbf{Dur}}: Duration, \textit{\textbf{Rep}}: Repetition, \textit{\textbf{Temp Topo}}: Temporal Topology, \textit{\textbf{Causal Cont}}: Causal Contingency, and \textit{\textbf{Cross Comp}}: Cross-Event Comparison. We include the overall performance (\textit{\textbf{Avg}}) for each high-level category alongside the fine-grained categories. The results show clear performance gaps across categories. Models generally perform well on discrete \textit{Events}, but struggle with \textit{Event Attributes}, which require low-level physical understanding such as \textit{Speed} and \textit{Force}. The best results are \textbf{bolded}.}

\label{tab:category}
\begin{tabularx}{\textwidth}{@{}>{\raggedright\arraybackslash}X
c c >{\columncolor{gray!15}}c
c c c c c c >{\columncolor{gray!15}}c
c c c >{\columncolor{gray!15}}c} %
\toprule
\multirow{2}{*}{\textbf{Models}} &
\multicolumn{3}{c}{\textbf{Event}} &
\multicolumn{7}{c}{\textbf{Event Attribute}} &
\multicolumn{4}{c}{\textbf{Structural Event Logic}} \\
\cmidrule(lr){2-4} \cmidrule(lr){5-11} \cmidrule(lr){12-15}
& \makecell[c]{FG\\Action}
& \makecell[c]{State\\Trans}
& \textbf{Avg}
& Speed
& Force
& Mag
& Dir
& Dur
& Rep
& \textbf{Avg.}
& \makecell[c]{Temp\\Topo}
& \makecell[c]{Causal\\Cont}
& \makecell[c]{Cross\\Comp}
& \textbf{Avg} \\
\midrule

Random Chance
& 6.3 & 6.3 & 6.3
& 6.3 & 6.3 & 6.3 & 6.3 & 6.3 & 6.3 & 6.3
& 6.3 & 6.3 & 6.3 & 6.3 \\
\midrule

\multicolumn{15}{c}{\textbf{\textit{Open-Source Models}}} \\
\midrule

Qwen3-VL-235B
& 39.8 & 49.1 & 45.1
& 3.6 & 5.6 & 8.3 & 20.0 & 10.7 & 13.9 & 12.7
& \textbf{21.3} & 7.5 & \textbf{25.0} & \textbf{19.3} \\
Molmo2-8B
& \textbf{56.8} & \textbf{54.3} & \textbf{55.4}
& \textbf{10.7} & \textbf{11.1} & \textbf{12.5}
& \textbf{27.1} & \textbf{14.3} & \textbf{19.4} & \textbf{18.6}
& 20.4 & \textbf{15.0} & 18.2 & 18.8 \\
\addlinespace[2pt]
\midrule
\multicolumn{15}{c}{\textbf{\textit{Closed-Source Models}}} \\
\midrule
GPT-5
& \textbf{62.5} & \textbf{55.2} & \textbf{58.3}
& 21.4 & 16.7 & 25.0 & \textbf{40.0} & 42.9 & \textbf{30.6} & 32.4
& 49.1 & \textbf{50.0} & 45.5 & 48.4 \\
Gemini 3 Pro
& 50.0 & 49.1 & 49.5
& \textbf{57.1} & \textbf{50.0} & \textbf{45.8} & 24.3 & \textbf{46.4} & 27.8 & \textbf{37.3}
& \textbf{69.4} & 37.5 & \textbf{50.0} & \textbf{58.3} \\

\bottomrule
\end{tabularx}
\end{table*}

\subsection{Additional Category-Wise Diagnosis}
\label{sec:cat_results_more}
To pinpoint specific cognitive deficits, we analyze performance across the TimeBlind taxonomy in Table~\ref{tab:category}. 
\begin{itemize}[leftmargin=*,itemsep=2pt,topsep=0pt,partopsep=0pt,parsep=0pt]
    \item \textbf{Performance Across Hierarchical Levels. } We observe notable performance gaps across different temporal understanding tasks, with models generally performing better on discrete \emph{Events} (e.g., distinguishing atomic actions) compared to continuous \emph{Event Attributes} or \emph{Structural Event Logic}. For example, GPT-5 achieves 58.3\% accuracy in the \emph{Event} category (peaking at 62.5\% for \emph{Fine-Grained Action}), but performance declines to 32.4\% for \emph{Event Attributes} and 48.4\% for \emph{Structural Event Logic}. 
    \item \textbf{Physical Dynamics.} Performance is lowest in the \emph{Event Attributes} category, which requires sensitivity to continuous or qualitative parameters such as kinematics (speed) and dynamics (force, magnitude). Strong proprietary models GPT-5 and Gemini 3 Pro only achieve 32.4\% and 37.3\% I-Acc, respectively, while the top-performing open-source model, Molmo2-8B, scores only 18.6\%, struggling to distinguish nuances such as \textit{gentle vs. forceful}, \textit{fast vs. slow}, or \textit{large vs. small amplitude}. These results expose a systematic deficiency in current models' understanding of low-level, physics-related temporal dynamics.
    \item \textbf{Gap in Structural Logic.} We observe that GPT-5 and Gemini 3 Pro demonstrate relatively robust performance in \emph{Structural Logic Analysis}, achieving 48.4\% and 58.3\%, respectively. However, a striking disparity emerges between these proprietary models and their open-source counterparts. The strong open-source model Qwen3-VL-235B achieves only 19.3\%—trailing Gemini 3 Pro by 39.0\%—and scores a mere 7.5\% in the \emph{Causal Contingency} sub-category. These results indicate that while open-source MLLMs may effectively detect isolated events, they lack the necessary logical abilities to reason about precise relationships between multiple events, particularly regarding causality.
\end{itemize}

\subsection{Additional Ablation Studies}
\label{sec:more_ablation_study}

\noindent\textbf{Video Similarity.} We computed CLIP-based video similarity for all video pairs by sampling frames at 1 FPS, extracting frame-level embeddings, and average-pooling them within each video. We further evaluated GPT-5 on two similarity-based subsets, namely the top 30\% highest-similarity pairs and the bottom 30\% lowest-similarity pairs. As shown in Table~\ref{tab:similarity_subset}, GPT-5's performance is comparable across the highest- and lowest-similarity subsets (45.6\% vs 47.8\%), indicating that the benchmark's difficulty does not depend strongly on residual visual differences between paired videos. Combined with the shortcut analysis in Table~\ref{tab:temporal_tests} (single-frame, shuffled-frame, and language-only baselines all near random chance), this confirms that TimeBlind is robust to subtle residual visual cues. 

\begin{table}[h!]
\centering
\small
\caption{\textbf{GPT-5 Performance on different similarity-based subsets.} The table reports GPT-5's I-Acc results on the high-similarity subset, the low-similarity subset, and the full benchmark. GPT-5's performance is comparable across the highest- and lowest-similarity subsets (45.6\% vs 47.8\%), indicating that the benchmark's difficulty does not depend strongly on residual visual differences between paired videos.}
\label{tab:similarity_subset}
\begin{tabular}{l c}
\toprule
\textbf{Subset} & \textbf{I-Acc} \\
\midrule
Top 30\% (highest similarity)   & 45.6 \\
Bottom 30\% (lowest similarity) & 47.8 \\
Full benchmark                  & 46.3 \\
\bottomrule
\end{tabular}
\vspace{-5pt}
\end{table}

\noindent\textbf{Video Source Heterogeneity.}
We further analyze model performance across the three video sources: Internet Retrieval, Human Recording, and Simulation. We report per-source I-Acc for representative models in Table~\ref{tab:source_heterogeneity}. As the results show, Gemini 3 Pro performs consistently across all three sources, while GPT-5 and Qwen3-VL-235B show a clear drop on Simulation data. Moreover, Qwen3-VL-235B falls well behind the leading closed-source models on all three sources---for example, it trails GPT-5 by 20.8\% on Internet Retrieval. These results suggest that our benchmark's difficulty is not driven by any single video source.

\begin{table}[t]
\centering
\small
\caption{\textbf{Model performance across video sources.} We report I-Acc (\%) on the three video sources: Internet Retrieval, Human Recording, and Simulation. Gemini 3 Pro performs consistently across sources, while GPT-5 and Qwen3-VL-235B degrade on Simulation data.}
\label{tab:source_heterogeneity}
\setlength{\tabcolsep}{5pt}
\begin{tabular}{lccc}
\toprule
\textbf{Model} & \textbf{Internet} & \textbf{Recording} & \textbf{Simulation} \\
\midrule
Qwen3-VL-235B & 28.5 & 26.9 & 19.1 \\
GPT-5         & 49.3 & 47.4 & 39.1 \\
Gemini 3 Pro  & 47.9 & \textbf{48.6} & \textbf{47.3} \\
\bottomrule
\end{tabular}
\vspace{-5pt}
\end{table}

\subsection{Failure Cases}
\label{app_sec:failure_cases}
\begin{figure*}[h]
  \centering
  \includegraphics[width=0.8\textwidth]{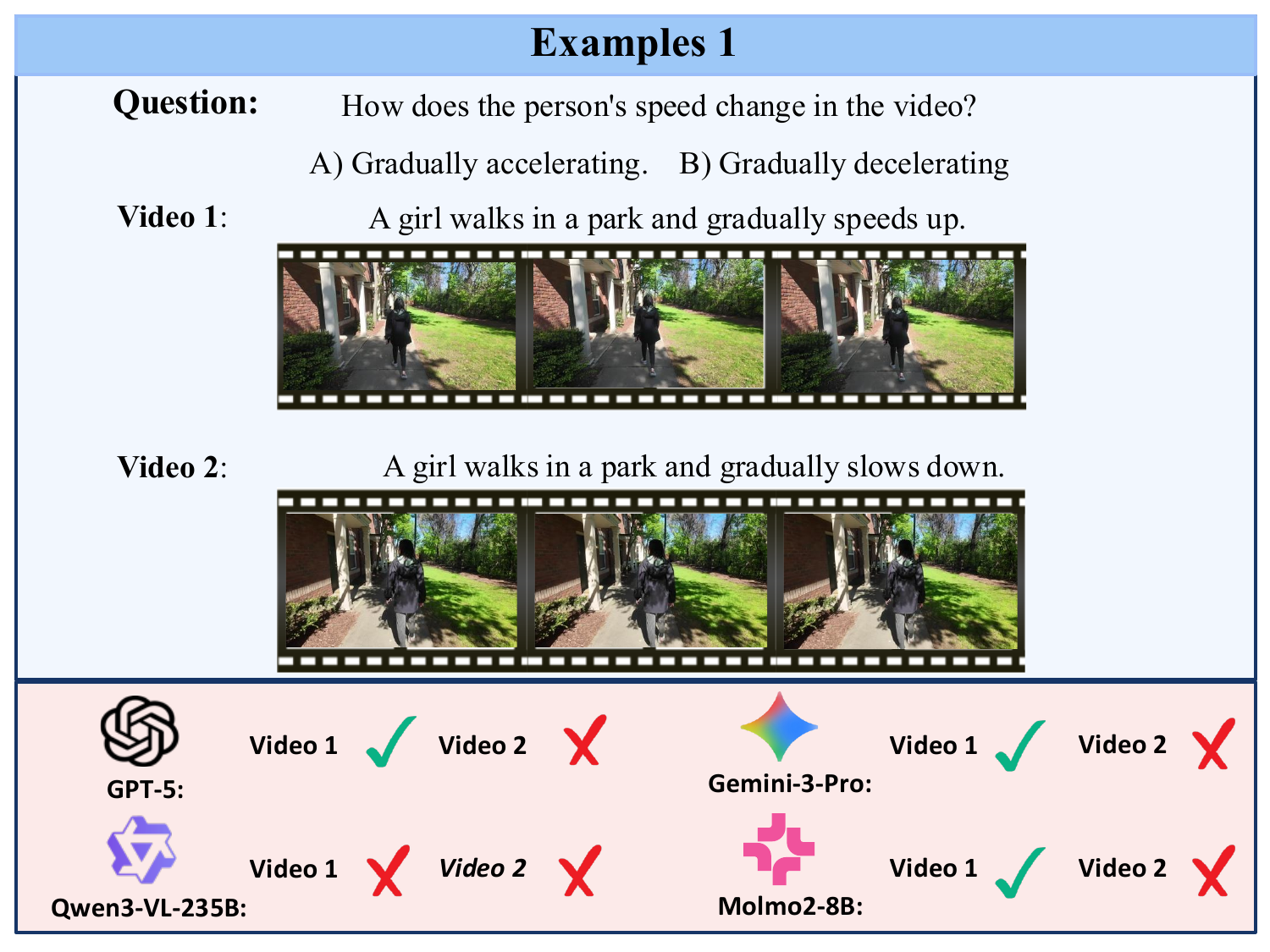}
  \caption{Failure Case 1}
  \label{fig:failure_case_1}
\end{figure*}

\begin{figure*}[h]
  \centering
  \includegraphics[width=0.8\textwidth]{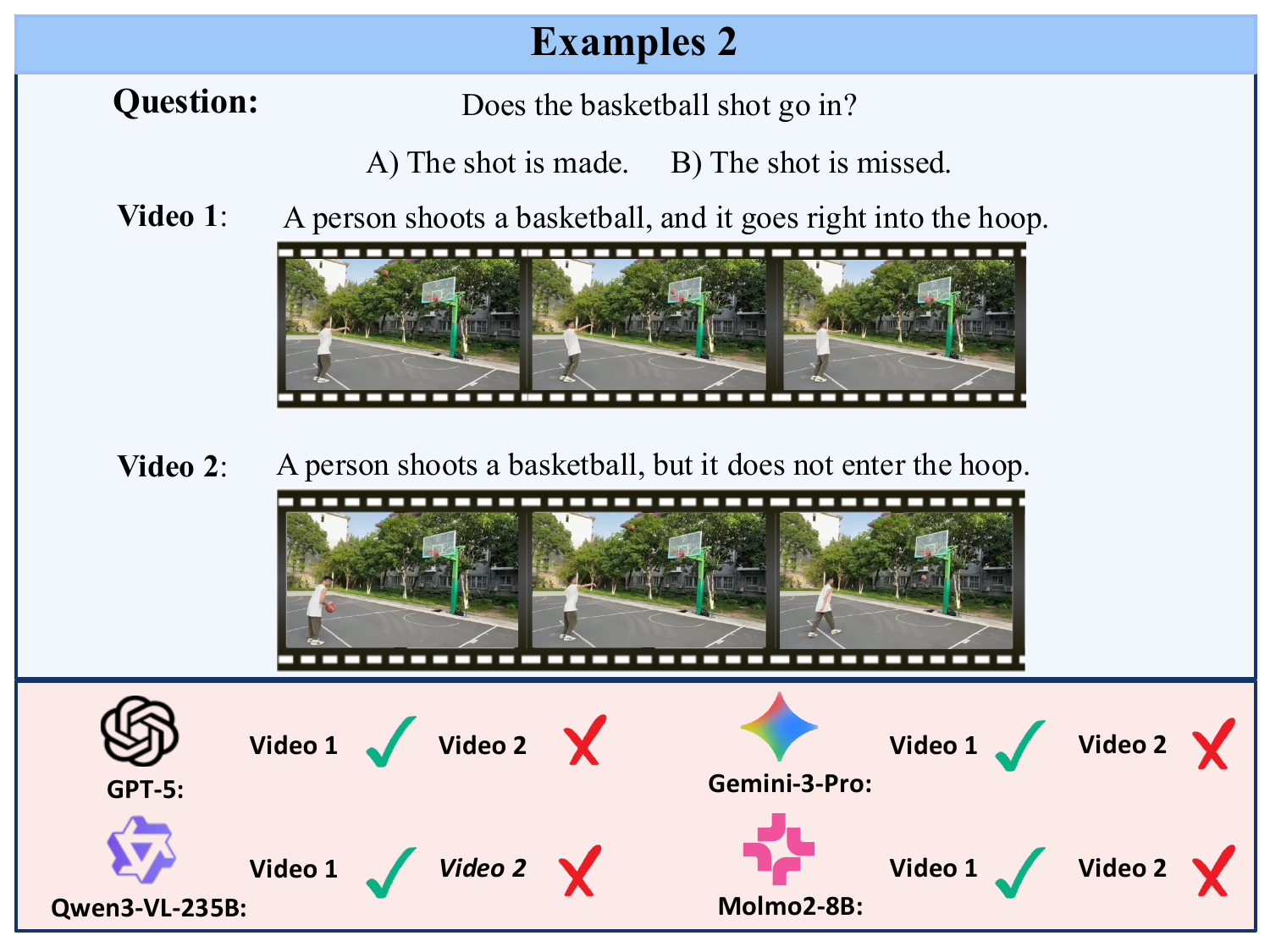}
  \caption{Failure Case 2}
  \label{fig:failure_case_2}
\end{figure*}

\begin{figure*}[h]
  \centering
  \includegraphics[width=0.8\textwidth]{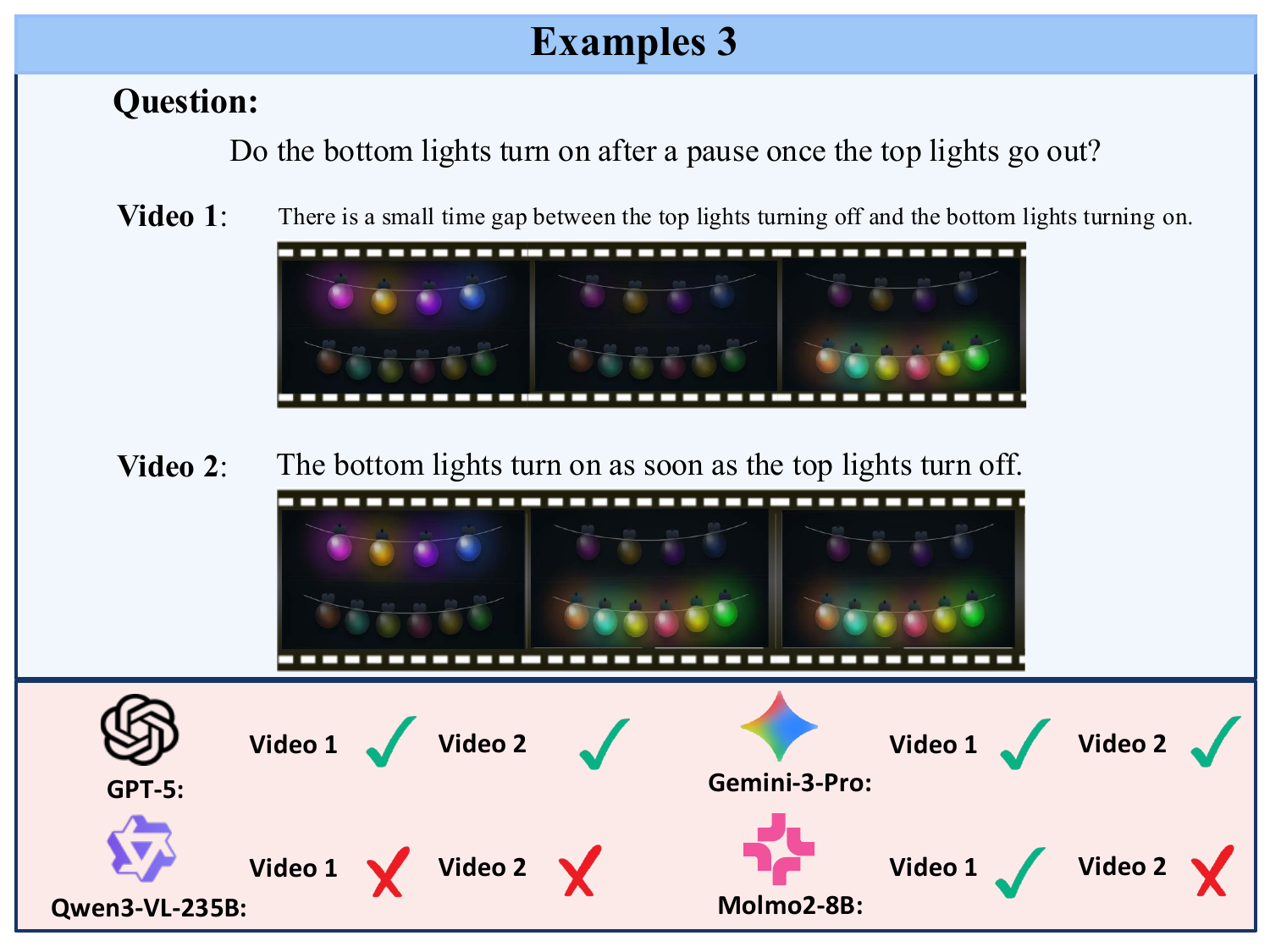}
  \caption{Failure Case 3}
  \label{fig:failure_case_3}
\end{figure*}

\begin{figure*}[h]
  \centering
  \includegraphics[width=0.8\textwidth]{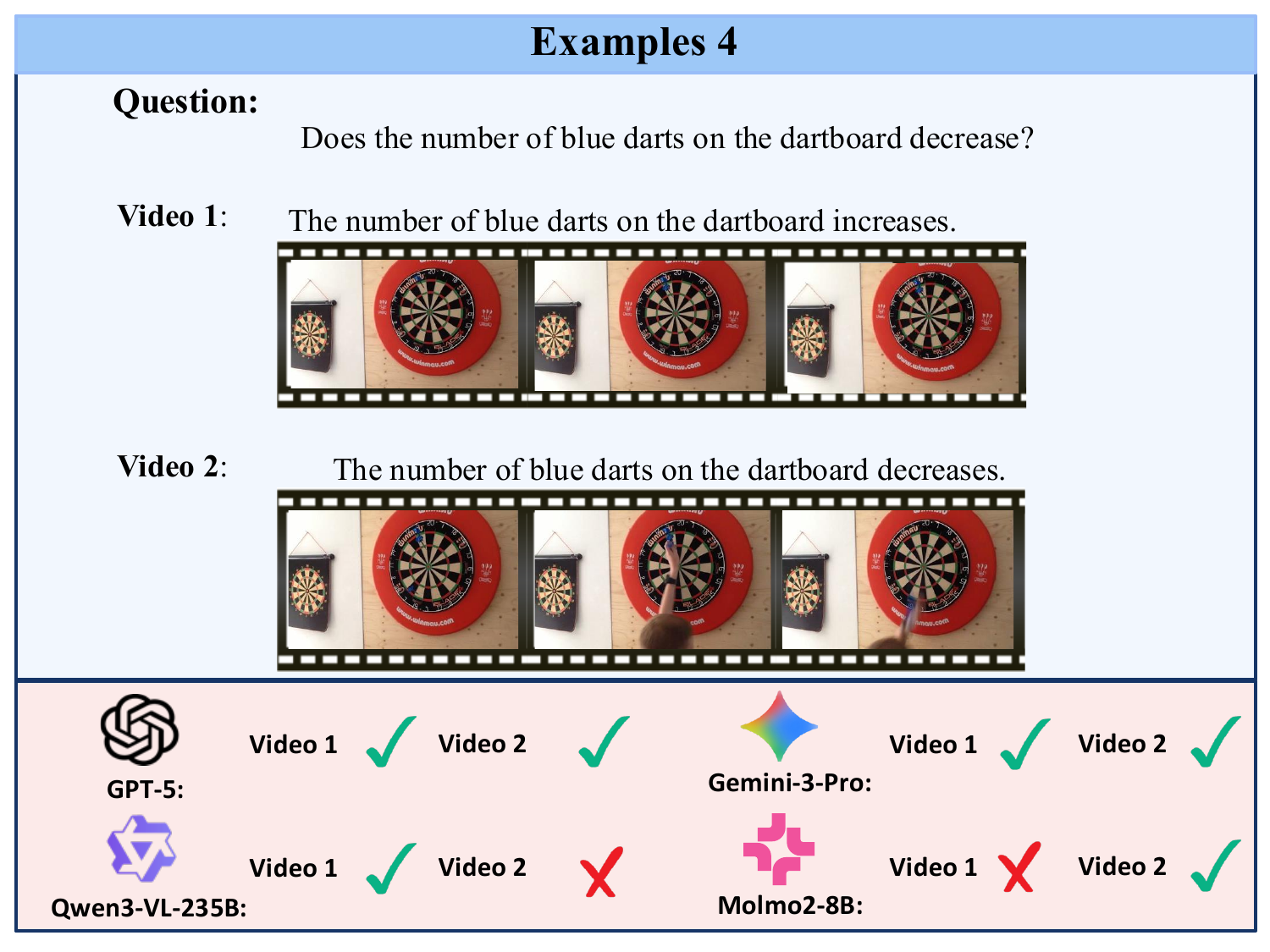}
  \caption{Failure Case 4}
  \label{fig:failure_case_4}
\end{figure*}

\begin{figure*}[h]
  \centering
  \includegraphics[width=0.8\textwidth]{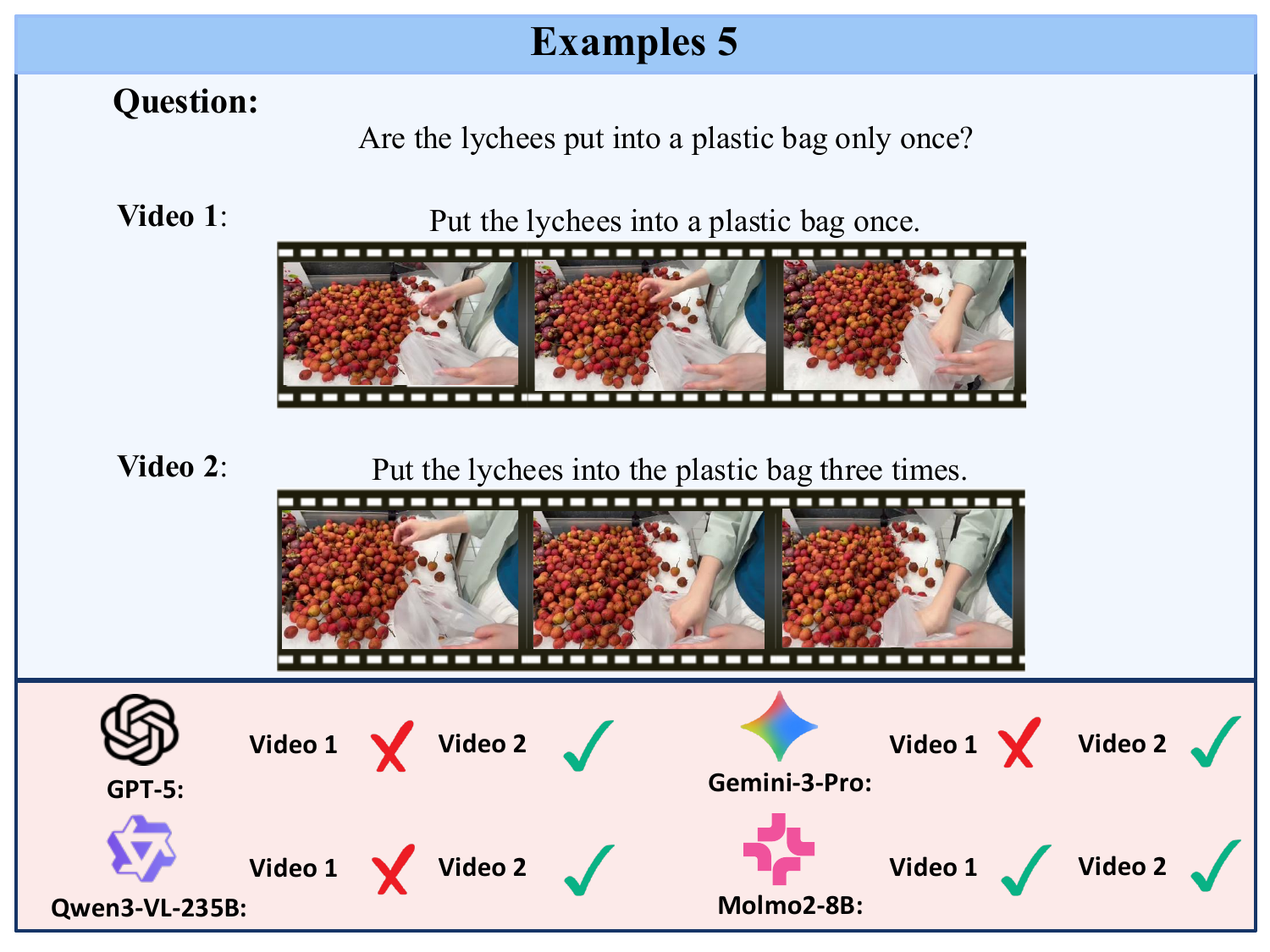}
  \caption{Failure Case 5}
  \label{fig:failure_case_5}
\end{figure*}

\begin{figure*}[h]
  \centering
  \includegraphics[width=0.8\textwidth]{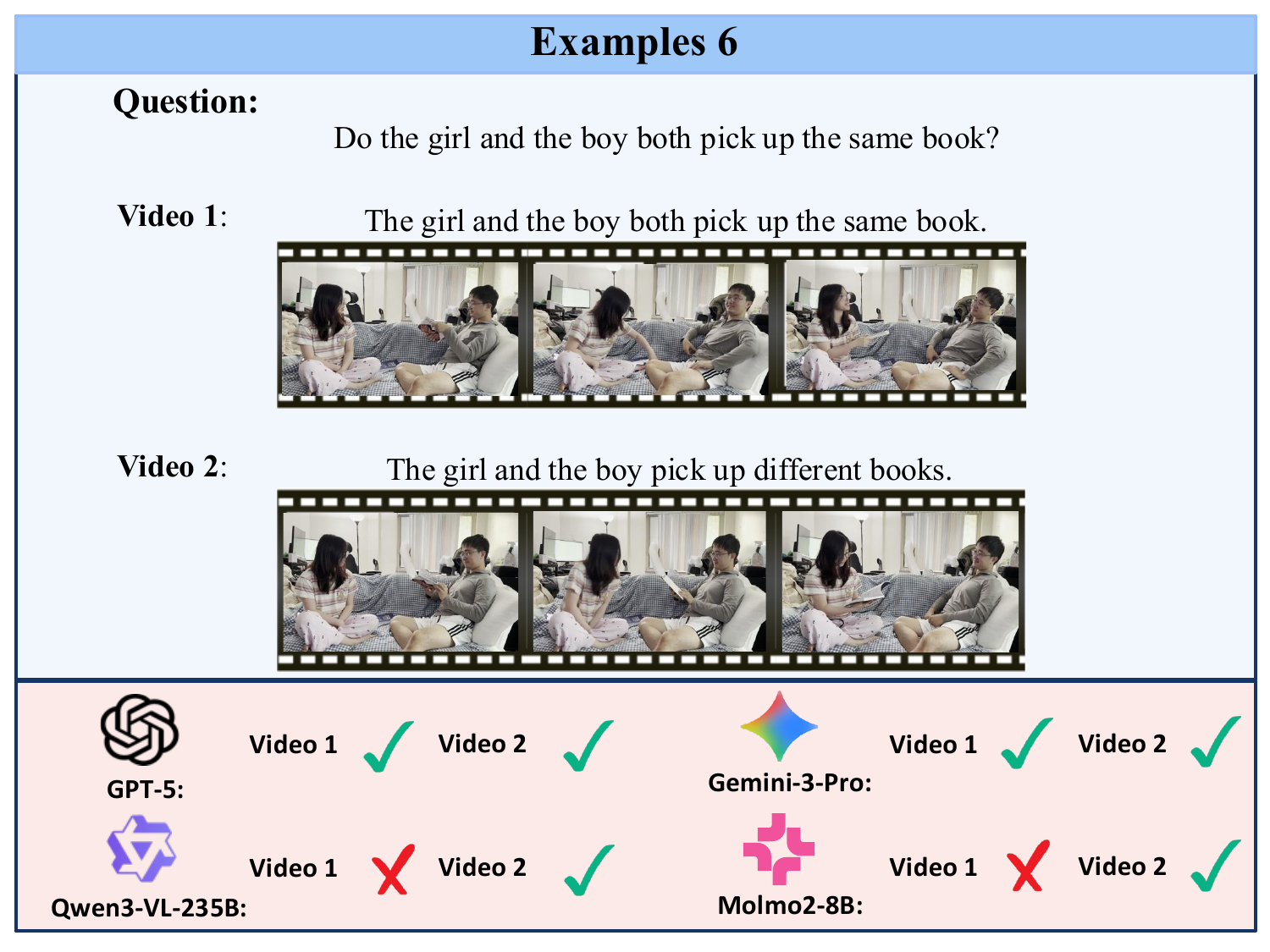}
  \caption{Failure Case 6}
  \label{fig:failure_case_6}
\end{figure*}

\begin{figure*}[h]
  \centering
  \includegraphics[width=0.8\textwidth]{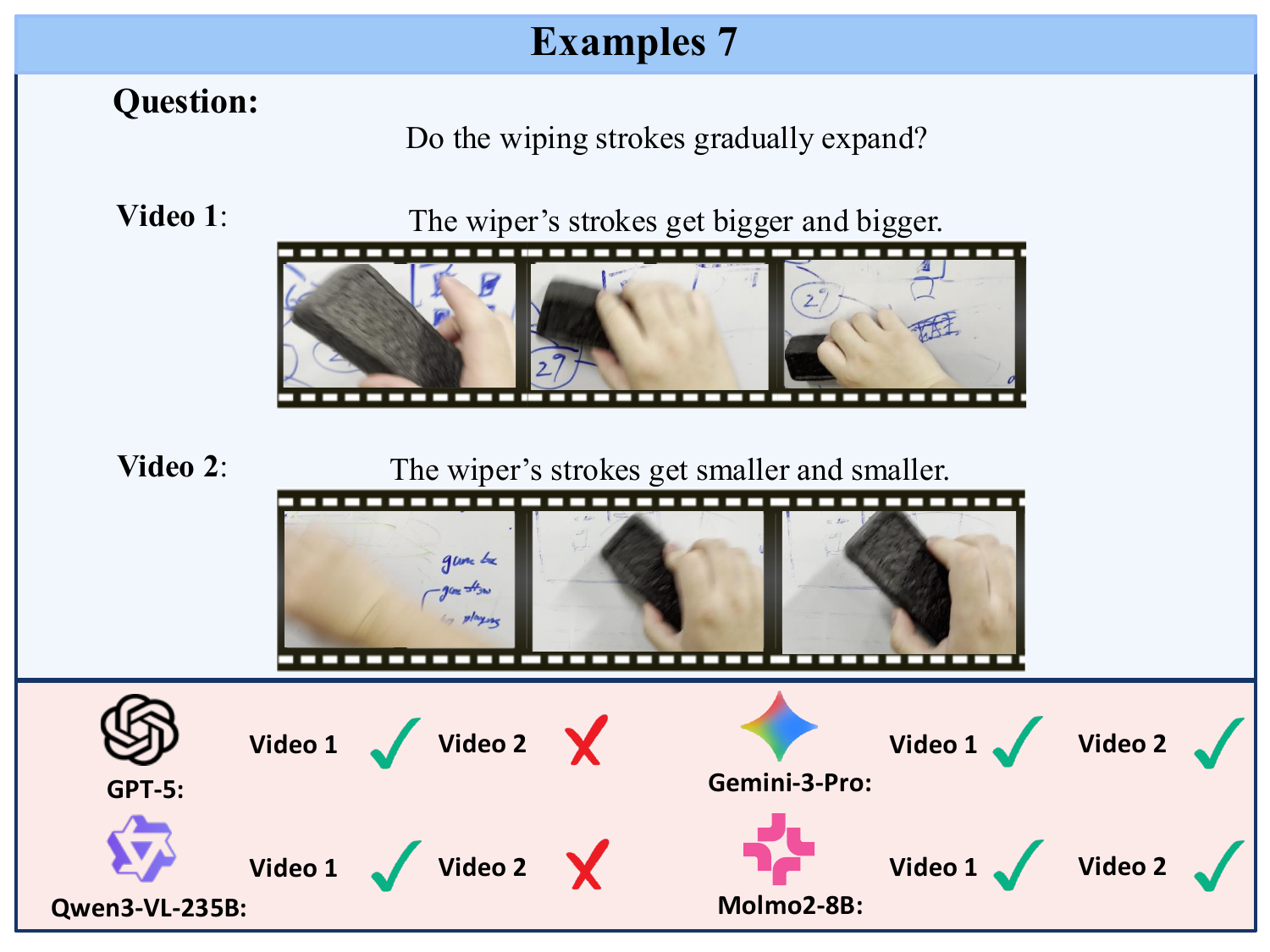}
  \caption{Failure Case 7}
  \label{fig:failure_case_7}
\end{figure*}

In Figures~\ref{fig:failure_case_1}--\ref{fig:failure_case_7}, we present representative failure cases of state-of-the-art models, covering both closed-source (GPT-5 and Gemini 3 Pro) and open-source models (Qwen3-VL-235B and Molmo2-8B). Each figure shows one question, the two paired videos with their corresponding captions, and each model's answer to the question on both videos, with check marks and cross marks indicating correct and incorrect answers. These examples illustrate common error patterns, such as failing to distinguish speed changes (Figure~\ref{fig:failure_case_1}), miscounting repetitions (Figure~\ref{fig:failure_case_5}), and misjudging temporal relations between events (Figure~\ref{fig:failure_case_3}).

\subsection{Impact Statement} 
\label{sec:impact}
This work introduces a diagnostic benchmark for evaluating temporal understanding in video-language models. We anticipate several positive impacts: (1) TimeBlind can guide development of more temporally-aware models, which is critical for applications in robotics, autonomous driving, and assistive technologies; (2) by revealing systematic failure modes, our benchmark may help practitioners avoid deploying models in safety-critical scenarios where temporal reasoning is essential.

We also acknowledge potential concerns. Improved video understanding capabilities could enable more sophisticated surveillance or content moderation systems, raising privacy considerations. Additionally, our benchmark primarily features videos from controlled settings and internet sources, which may not represent the full diversity of real-world scenarios or populations. We encourage future work to expand temporal reasoning evaluation to more diverse contexts.

The benchmark itself does not enable direct harmful applications, as it evaluates rather than enhances model capabilities. We will release our data and code to support reproducible research.

\subsection{License and Intended Use}
\label{sec:license}

\noindent \textbf{License.} We release the TimeBlind benchmark—including all curated video–question pairs, annotations, and evaluation code—under the Creative Commons Attribution 4.0 International License (CC BY 4.0)\footnote{\url{https://creativecommons.org/licenses/by/4.0/}}. For videos sourced from the internet, we retain the original license terms of each source. For videos recorded by participants and generated via simulation, we release them under CC BY 4.0.

\noindent \textbf{Intended Use.} TimeBlind is intended solely as a diagnostic research benchmark for evaluating the compositional spatio-temporal understanding capabilities of multimodal large language models. It is not designed for training, fine-tuning, or deployment in production systems, and should not be used as a basis for safety-critical decisions involving real-world video analysis.

\noindent \textbf{Use of Existing Artifacts.} Our use of existing artifacts is consistent with their intended use. We evaluate publicly released MLLMs (e.g., GPT-5, Gemini 3 Pro, Qwen3-VL) in zero-shot inference settings consistent with their official APIs and documented evaluation protocols. Internet-sourced videos were collected from publicly accessible platforms (e.g., YouTube) for non-commercial research purposes, in line with platform terms of service. Synthetic videos generated via simulation tools (e.g., Unity) are produced and distributed in accordance with the tools' research-use terms. During the coding process, we used Claude Code as an AI-assisted programming tool.

\vspace{1.0cm}

\end{document}